%% file: main.tex
\documentclass[runningheads]{llncs}
\pdfoutput=1
\usepackage[pdftex]{graphicx}
\DeclareGraphicsExtensions{.pdf,.jpeg,.png}

\usepackage{eccv}

\usepackage{eccvabbrv}

\usepackage{graphicx}
\usepackage{booktabs}
\usepackage{lipsum}
\usepackage{mathtools}
\usepackage{amsmath}
\usepackage{amssymb}
\usepackage{multirow}
\usepackage{float}
\usepackage{wrapfig}
\usepackage{bm}
\usepackage{grffile}
\renewcommand{\paragraph}[1]{\vspace{-1.5mm}{\flushleft\textbf{#1}}} 

\usepackage[accsupp]{axessibility}  

\usepackage[pagebackref=false,breaklinks=true,colorlinks=true,citecolor=blue,bookmarks=false]{hyperref}

\usepackage{orcidlink}

\begin{document}

\title{JointDreamer: Ensuring Geometry Consistency and Text Congruence in Text-to-3D Generation via Joint Score Distillation} 

\titlerunning{JointDreamer}

\author{Chenhan Jiang\inst{1}*$^\dagger$\orcidlink{0000-0001-8771-3641} \and
Yihan Zeng\inst{2}* \and
Tianyang Hu\inst{2} \and
Songcun Xu\inst{2} \and
Wei Zhang\inst{2} \and
Hang Xu\inst{2} \and
Dit-Yan Yeung\inst{1}}

\authorrunning{C.~Jiang et al.}

\institute{The Hong Kong University of Science and Technology \and
Huawei Noah's Ark Lab \\
$^*$Equal contribution.  \hspace{2mm}$^\dagger$Corresponding Author: \email{jchcyan@gmail.com}\\
\url{https://jointdreamer.github.io}
}

\maketitle

\input{tex/0_abstract}  
\input{tex/1_intro}

\input{tex/2_related}

\input{tex/3_pre}
\input{tex/3_method}
\input{tex/4_experiment}
\input{tex/5_conclusion}

%
%
\bibliographystyle{splncs04}
\bibliography{main}

\input{sub_tex/supp}

\end{document}

%% file: tex/0_abstract.tex
\begin{abstract}
Score Distillation Sampling (SDS) by well-trained 2D diffusion models has shown great promise in text-to-3D generation. However, this paradigm distills view-agnostic 2D image distributions into the rendering distribution of 3D representation for each view independently, overlooking the coherence across views and yielding 3D inconsistency in generations.
In this work, we propose \textbf{J}oint \textbf{S}core \textbf{D}istillation (JSD), a new paradigm that ensures coherent 3D generations. 
Specifically, we model the joint image distribution, which introduces an energy function to capture the coherence among denoised images from the diffusion model. We then derive the joint score distillation on multiple rendered views of the 3D representation, as opposed to a single view in SDS.
In addition, we instantiate three universal view-aware models as energy functions, demonstrating compatibility with JSD.
Empirically, JSD significantly mitigates the 3D inconsistency problem in SDS, while maintaining text congruence. Moreover, we introduce the Geometry Fading scheme and Classifier-Free Guidance (CFG) Switching strategy to enhance generative details.
Our framework, JointDreamer, establishes a new benchmark in text-to-3D generation, achieving outstanding results with an 88.5\% CLIP R-Precision and 27.7\% CLIP Score. These metrics demonstrate exceptional text congruence, as well as remarkable geometric consistency and texture fidelity. 
\keywords{3D Vision \and 3D Generation \and Energy Function}
\end{abstract}

%% file: tex/1_intro.tex
\begin{figure}[t]
\hsize=\textwidth 
  \centering
  \includegraphics[width=0.98\textwidth]{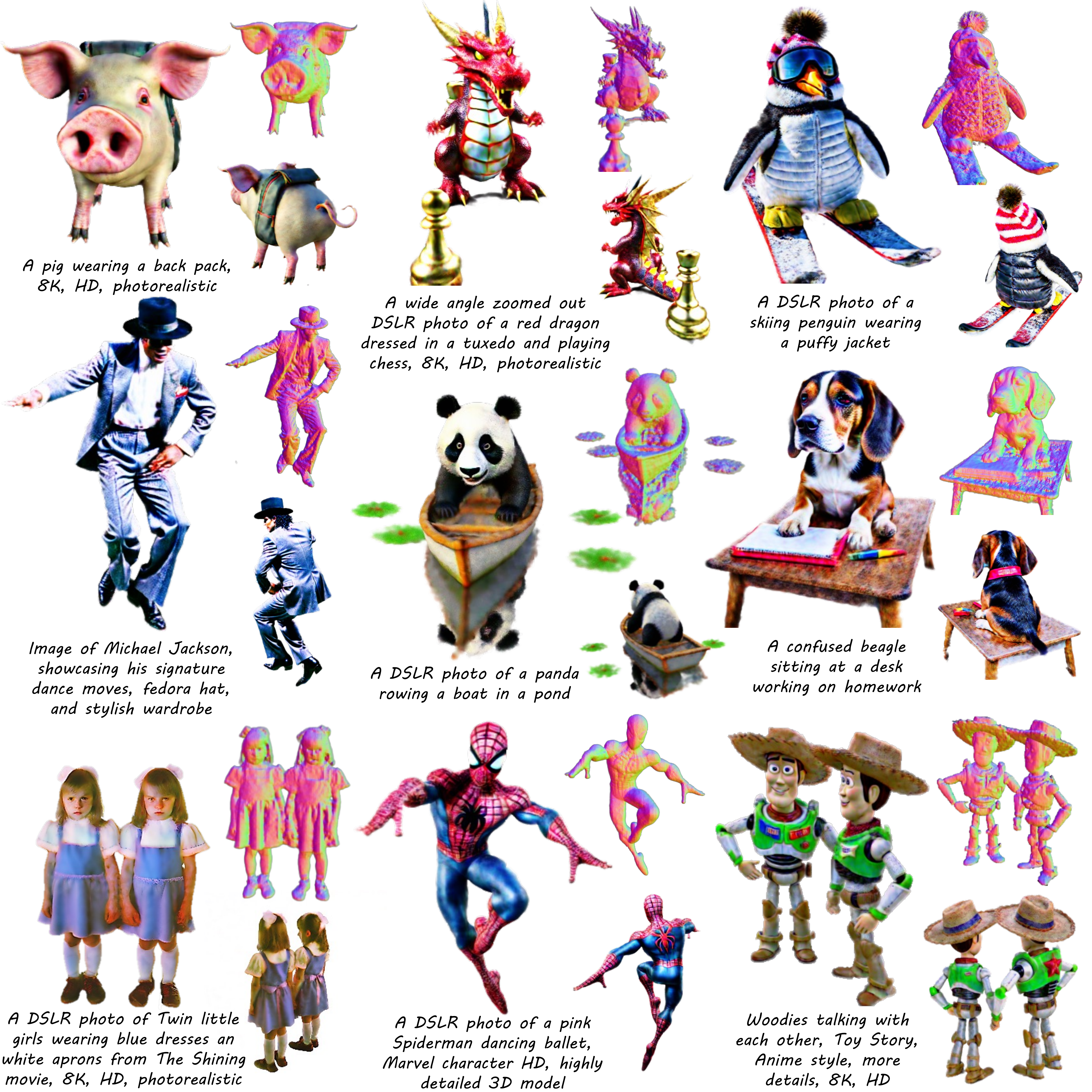}
   \vspace{-2mm}
   \caption{\textbf{{Text-to-3D generations by JointDreamer from scratch.}} JointDreamer excels in generating geometrically consistent and high-fidelity 3D assets, adhering to complex textual descriptions that are challenging for previous methods. 
	}
\label{fig:intro}
\vspace{-6mm}
\end{figure}

\section{Introduction}
\label{sec:intro}

3D content creation is essential for diverse applications, including gaming, robotics simulation, and virtual reality. 
However, it is labor-intensive, demanding substantial time for skilled designers to create a single 3D asset. Hence, automating 3D creation with text input has attracted considerable attention. Recently, the score distillation sampling (SDS) algorithm pioneered by DreamFusion~\cite{dreamfusion22} shows promise in text-to-3D tasks, which lifts image distribution from a well-trained diffusion model~\cite{sd2022} into parameterized 3D representation like NeRF~\cite{mildenhall2021nerf}.
Compared to 3D generative models~\cite{nichol2022pointe, jun2023shape, cao2023large} that struggle with producing arbitrary objects due to limited text-3D training data, SDS-based methods~\cite{dreamfusion22, magic3d22, wang2023prolificdreamer,chen2023fantasia3d} can generate arbitrary 3D assets with diverse text input.

\begin{wrapfigure}{r}{0.45\textwidth}
\vspace{-8mm}
  \centering
   \includegraphics[width=1\linewidth]{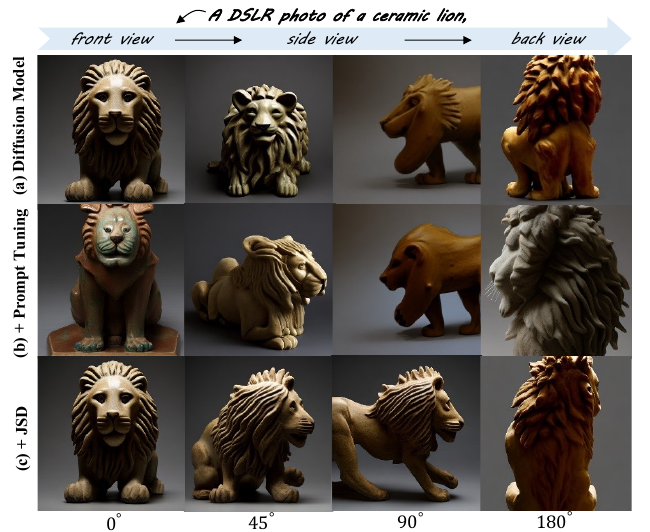}
   \vspace{-5mm}
   \caption{\textbf{Illustration of text-conditioned images for different viewpoints}, where input texts are augmented with corresponding direction prompts for each view. \textbf{(a)} The original generations from 2D diffusion model~\cite{sd2022} are view-agnostic and inconsistent across views. \textbf{(b)} Text prompt tuning~\cite{perpneg23} has limited improvement in the directional structure of generated images for each view. \textbf{(c)} JSD injects coherence measurement from the proposed binary classifier (refer to Section~\ref{sec:view}), contributing to modified directional structures and semantical consistency across views.}
   \label{fig:distribution}
   \vspace{-5mm}
\end{wrapfigure}

Although SDS-based methods benefit from the generalizability of diffusion models, they often encounter a common issue known as Janus artifacts ~\cite{dreamfusion22, magic3d22, wang2023prolificdreamer,chen2023fantasia3d}.
These artifacts manifest as repeated content from different viewpoints of a 3D generation, yielding a lack of realism and coherence in the rendered views.
We investigate the Janus artifacts by visualizing image distributions of the diffusion model~\cite{sd2022} from multiple viewpoints, as illustrated in Fig.~\ref{fig:distribution}\textcolor{red}{(a)}. The results reveal the view-agnostic nature and the content inconsistency across views of the diffusion model. 
Consequently, SDS optimizes each rendered view of 3D representation independently and inherits image distribution without coherent multi-view perspective, leading to the geometric inconsistency of 3D generations.

Existing works~\cite{perpneg23,li2023instant3d} address the aforementioned challenges within the SDS framework by employing prompt engineering techniques. However, the effectiveness of such methods remains unsatisfactory, as evident from the results depicted in Fig.~\ref{fig:distribution}\textcolor{red}{(b)}. 
Alternative methods~\cite{shi2023mvdream,hu2024efficientdreamer} propose to finetune view-aware diffusion models using rendered images of 3D datasets~\cite{deitke2023objaverse,deitke2023objaversexl}. 
Nevertheless, they are prone to overfitting on limited text-3D training data~\cite{li2023sweetdreamer}, decreasing the text congruence when handling complex text inputs. 
Based on the above observations, it requires a rethinking of the SDS optimization to enhance the 3D coherence of generations while maintaining generalizability.

In this work, we first present Joint Score Distillation (JSD), which significantly promotes the 3D consistency of generation and inherits generalizability from diffusion models.
Specifically, we model the joint image distribution of diffusion model via an energy function measuring coherence across denoised images.
It facilitates the extension of the KL-Divergence in SDS from single-view into multi-view. We then derive the joint score distillation function from multi-view KL-Divergence, which ensures inter-view coherence in the optimization process of 3D generation.
We show that SDS is a special case of JSD with the energy term omitted, which indicates the absence of coherence constraint across views.

Building upon JSD optimization, we present three view-aware models as energy terms to showcase the compatibility of JSD: the Binary Classification Model, the Image-to-Image Translation Model, and the Multi-view Generation Model. Through empirical analysis, it is observed that different view-aware models introduce distinct coherence measurements, leading to diverse 3D generations, while all contributing to 3D consistency. Furthermore, to facilitate a more comprehensive comparison with existing text-to-3D generation methods, we introduce JointDreamer, an innovative framework capable of producing geometric-consistent and high-fidelity 3D assets adhering to complex text descriptions. Notably, in addition to incorporating a Multi-view Generation model as an energy term in JSD, we introduce two complementary techniques, namely the Geometry Fading scheme and the Classifier-Free Guidance (CFG) Switching strategy, to enhance generative details.

We systematically assess the quality of our approach, both qualitatively and quantitatively, compared to existing methods. Qualitative results gallery can be found in Fig.~\ref{fig:intro}. Our JointDreamer consistently produces high-fidelity 3D assets and mitigates Janus artifacts in SDS. It maintains text congruence even when confronted with complex text input, achieving 88.5\% CLIP R-Precision and 27.7\% CLIP Score. 

In brief, our contributions are summarized as follows:
\begin{itemize}
    \item We introduce a novel Joint Score Distillation (JSD) for text-to-3D generation, optimizing multiple views jointly via an energy function to capture inter-view coherence.
    \item We present three view-aware models as energy functions to show compatibility with JSD, all of them mitigate the Janus problem in SDS.
    \item We introduce the text-to-3D framework JointDreamer, incorporating complementary Geometry Fading and CFG Switching techniques. Our JointDreamer achieves geometrically consistent and high-fidelity 3D assets even with complex textual inputs.
\end{itemize}

%% file: tex/2_related.tex
\section{Related Works}
\label{sec:related}
\paragraph{Text-to-3D Generation.}
Existing text-to-3D generation methods can be categorized into two streams: 3D generative models and 2D optimization methods. The former encompasses various deep generative models such as Variational Auto Encoders (VAEs)~\cite{henderson2020learning,henderson2020leveraging}, Generative Adversarial Models (GAN)~\cite{nguyen2019hologan, niemeyer2021giraffe, deng2022gram, gao2022get3d, chan2022efficient}, diffusion models~\cite{nichol2022pointe, chen2023single, liu2023meshdiffusion} and transformer architectures~\cite{cao2023large, jun2023shape}. These models are efficient in inference but often struggle with generalizability and training stability, attributed to the limited scope and complexity of available 3D datasets. The latter approach centers around the Score Distillation Sampling (SDS) algorithm proposed by~\cite{dreamfusion22}, which leverages 2D diffusion model priors\cite{sd2022} for optimizing 3D representations.  Subsequent advancements have refined this technique by improving the 3D representations~\cite{magic3d22, chen2023fantasia3d}, the sampling scheduler~\cite{huang2023dreamtime} and loss design~\cite{wang2023prolificdreamer}. However, the above approaches overlook the geometric consistency problem, facing inherent multi-face Janus issues in SDS. Prior works try to alleviate the Janus issues with prompt tuning~\cite{perpneg23} yet they achieve limited effect. Very recent work MVDream~\cite{shi2023mvdream} addresses the problem by fine-tuning a multi-view diffusion model, but it is susceptible to overfitting on scarce 3D training data, compromising semantic consistency in text-to-3D generations. In this work, we address the fundamental flaw of SDS that optimizes each view independently by introducing a  joint optimization function that enforces inter-view consistency, essentially solving the Janus issues in SDS while preserving its generalizability.

\paragraph{Diffusion-based Novel View Synthesis. }
As an alternative to 3D generation, novel view synthesis models the challenge as a view-conditioned image-to-image translation task. There have been proposals for pose-conditioned image-to-image diffusion models~\cite{watson2022novel} that generate novel views on synthetic data in 3D. Recently, ~\cite{liu2023zero} promoted the generalizability of novel view synthesis by fine-tuning the 2D diffusion model~\cite{sd2022} on renderings of 3D dataset, which facilitates images-to-3D tasks with 3D reconstruction or SDS algorithm. To further improve the 3D consistency across generated views and input view, very recent works~\cite{shi2023zero123++, long2023wonder3d, liu2023syncdreamer} modify the generation process into multi-view generation and present corresponding architecture designs. Generally, these methods take camera specifications as conditions and enable the viewpoint-aware generations, but they can hardly accurately capture a complete 3D scene consistently and densely. 
 Our method acknowledges the potential of these models to discern relative inter-view relationships, which we harness to provide inter-view coherence for our JSD, thus serving as universal guidance models.

%% file: tex/3_pre.tex
\begin{figure*}[t]
  \centering
   \includegraphics[width=1.0\linewidth]{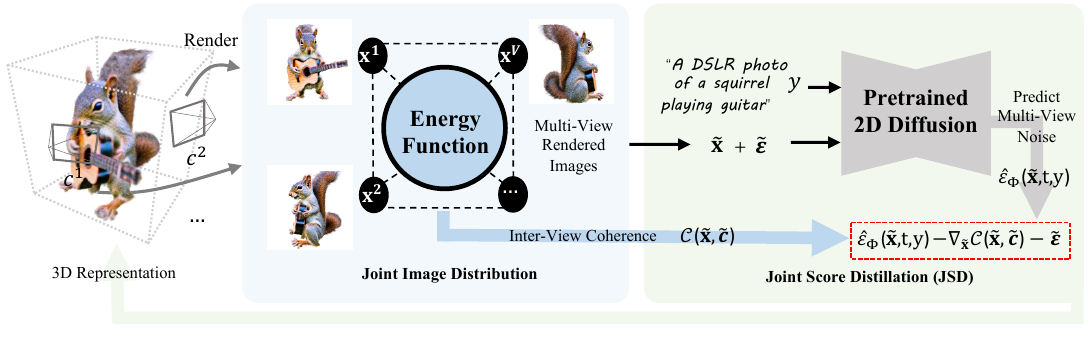}
\vspace{-8mm}
   \caption{\textbf{Overview of JointDreamer Framework}. 
   We introduce an energy function to model the joint distribution for multi-view denoised images from 2D diffusion model, facilitating the Joint Score Distillation (JSD) optimization for text-to-3D generation.
   \vspace{-3mm}
   }
   
   \label{fig:pipeline}
\end{figure*}
\section{Preliminaries}\label{sec:pre}
We review the original SDS and address its fundamental limitation: optimizing independently for each single view of 3D representation and distilling with the view-agnostic image distribution of diffusion model. It results in geometric inconsistency issues, dubbed as Multi-Face Janus Problem.

\paragraph{Score Distillation Sampling (SDS).}
SDS optimization is widely adopted by text-to-3D generation pipelines~\cite{dreamfusion22,magic3d22,chen2023fantasia3d,wang2023prolificdreamer,luo2023diff}. Given a 3D representation with learnable parameters $\theta$ and a pre-trained 2D diffusion model with noise prediction network $\epsilon_{\Phi}(\text{x}_t, t, y)$, SDS optimizes $\theta$ by minimizing the KL-divergence: 
\begin{equation}\small
    \min\limits_{\theta}D_{KL}(q_t^{\theta}(\textbf{x}_t|c, y) || p_t(\textbf{x}_t|y)).
\label{eq:KLSDS}
\end{equation}
Here, $p_t(\text{x}_t|y)$ is the image distribution sampled from diffusion model, $q_t^{\theta}(\textbf{x}_t|c, y)$ is the distribution of rendered image $\text{x}_t=g(\theta, c)$ with respect to camera pose $c$ at timestep $t$ of the forward diffusion process, where $g$ is the renderer. 
To solve Eq.~(\ref{eq:KLSDS}), the score distillation function is derived as:
\begin{equation}\small
\begin{split}
    \nabla_{\theta}L_{SDS}(\theta) & \triangleq \mathrm{E}_{t,\textbf{x}}[w(t)\frac{\sigma_t}{\alpha_t}\nabla_{\theta}KL(q_t^{\theta}(\textbf{x}_t|c, y)||p_t(\textbf{x}_t|y))]\\
    & \triangleq \mathrm{E}_{t,\epsilon_{\Phi}}[w(t)(\hat{\epsilon}_{\Phi}(\textbf{x}_t, t, y)-\epsilon)\frac{\delta g(\theta, c)}{\delta \theta}],
\end{split}
\end{equation}
where $\hat{\epsilon}_{\Phi}:= (1+s)\epsilon_{\Phi}(\text{x}_t, t, y) - s\epsilon_{\Phi}(\text{x}_t, t, \emptyset)$ is modification of predicted noise with classifier-free guidance (CFG) $s$, $w(t)$ is time-dependent weighting function. 

\noindent\textbf{Multi-Face Janus Problem.}
To achieve consistency, it is essential that the rendering distributions $q_0^{\theta}(\textbf{x}_0|c, y)$ adhere to text condition $y$ and image distribution $p_t(\text{x}_t|y)$ keep consistency across views with different poses. For the image distribution $p_0(\textbf{x}_0|y)$ of 2D diffusion model, the pose condition can be injected via input text with the corresponding directional prompt~\cite{dreamfusion22, wang2023prolificdreamer}.
As illustrated in Fig.~\ref{fig:distribution}\textcolor{red}{(a)}, the pre-trained 2D image distribution, trained on individual images, is view-agnostic and lacks identity consistency across views. Even with a text tuning mechanism~\cite{perpneg23} specifically designed for multi-view image generation, as shown in Fig.~\ref{fig:distribution}\textcolor{red}{(b)}, the above issues are far from resolved. Since SDS minimizes KL-divergence between the image distribution and rendering distribution independently for each rendered view, it can only inevitably inherit the 3D-awareness deficit of the 2D diffusion model, resulting in inconsistent 3D generation, which is commonly referred to as the Multi-face Janus Problem of SDS.

%% file: tex/3_method.tex
\section{Method}
In this section, we introduce JointDreamer, a novel text-to-3D generation framework as illustrated in Fig.~\ref{fig:pipeline}. We first present the derivation of Joint Score Distillation (JSD) in Sec.~\ref{sec:JSD}, which extends the single-view optimization in SDS into a multi-view KL-Divergence. 
Then we integrate universal view-aware models into JSD to show the compatibility of JSD in Sec.~\ref{sec:view}, where we instantiate three kinds of view-aware models to capture inter-view coherence.
Finally, we elaborate on the overall framework JointDreamer in Sec.~\ref{sec:dreamer}, where we integrate the multi-view generation model into JSD. We also propose a geometry fading scheme and CFG switching strategy to further enhance generative quality.

\subsection{Joint Score Distillation (JSD)}\label{sec:JSD}
To address the multi-face problem arising from SDS, we extend the score distillation from single-view to multi-view settings and promote inter-view coherence across 2D image distribution,
and thus derive our JSD optimization function.

\paragraph{Coherence Modeling for Joint Image Distribution.} As discussed above, the rendering distributions of the 3D representation should maintain 3D consistency across views $\tilde{\textbf{x}} = \{\textbf{x}^1, \textbf{x}^2, \ldots, \textbf{x}^V\}$ with respect to different poses $\tilde{\bf c} = \{c^1, c^2, \ldots, c^V\}$. 
However, for 2D pre-trained diffusion models, different views are generated independently. 
To ensure consistency, we propose modeling the joint image distribution of multiple views, denoted as $p_0(\tilde{\textbf{x}}|\tilde{\textbf{c}}, y)$, within the diffusion model.
Following the commonly adopted assumption in energy-based distribution modeling \cite{weese2001shape, lecun2006tutorial, zhao2018geometric}, we introduce an energy function that measures the inter-view coherence by $\mathcal{C}(\tilde{\textbf{x}}, \tilde{\textbf{c}}):\mathbb{R}^{Vd}\to \mathbb{R}$ and define:  
\begin{equation}\label{eq:energy}\small
    p_0(\tilde{\textbf{x}}|\tilde{\textbf{c}}, y) \propto \exp(\mathcal{C}(\tilde{\textbf{x}}, \tilde{\textbf{c}})) \prod_{i=1}^V p_0(\textbf{x}^i|c^i, y),
\end{equation}
where a larger $\mathcal{C}(\tilde{\textbf{x}},\tilde{\textbf{c}})$ indicates greater coherence among the denoised view images. As a result, the joint image distribution is no longer view-independent.
In practice, the joint energy function can be implemented via various view-aware models, as long as they can reflect the coherence across multiple views. In Sec. \ref{sec:view}, we explore different choices in depth.
The modeling of coherence across joint image distributions facilitates our JSD on multi-view, integrating inter-view coherence to ensure consistent 3D representation.

\paragraph{KL-Divergence on Multiple Views.}
We extend the single-view KL-divergence in SDS to a multi-view version, based on the joint image distribution:
\begin{equation}\small
\label{eq:KLJSD}
\begin{split}
     & \min\limits_{\theta}D_{KL}(q_t^{\theta}(\tilde{\textbf{x}}|\tilde{\textbf{c}}, y) || p_t(\tilde{\textbf{x}}|\tilde{\textbf{c}}, y)) \\
     & = \min\limits_{\theta}D_{KL}(q_t^{\theta}(\tilde{\textbf{x}}|\tilde{\textbf{c}}, y) || \exp(\mathcal{C}(\tilde{\textbf{x}}, \tilde{\textbf{c}})) \prod_{i=1}^V p_t(\textbf{x}^i|c^i, y),
\end{split}
\end{equation} 
where the extra energy term $\mathcal{C}(\tilde{\bf x},\tilde{\bf c})$ in Eq.~\eqref{eq:energy} accounts for the inter-view coherence.
Without this constraint, e.g., $\mathcal{C}(\tilde{\bf x},\tilde{\bf c})\equiv 0$, different rendering views are optimized independently with the 2D diffusion model separately. 
In this sense, the original SDS can be seen as a special case of JSD. 

\paragraph{Joint Score Distillation Function.}
To correspond to the gradient of the new rule of multi-view KL-divergence in Eq.~\eqref{eq:KLJSD}, we derive our score distillation function that is jointly conducted on multiple views as follows:
\begin{equation}\label{eq:JSD}\small
\begin{split}
    &\nabla_{\theta}L_{JSD}(\theta) \\
    &\triangleq \mathrm{E}_{t,\epsilon_{\Phi}}[w(t)\mathbb{E}(\nabla_{\tilde{\textbf{x}}}\log q_t(\tilde{\textbf{x}}_t|\tilde{\textbf{x}}_0) -\nabla_{\tilde{\textbf{x}}}\log p_t(\tilde{\textbf{x}}_t|y))]\\ 
    & = \sum_{i=1}^V\mathrm{E}_{t,\epsilon^i_{\Phi}}[w(t)(\hat{\epsilon}_{\Phi}(\textbf{x}_{t}^{i}, y)
    -\frac{\partial \mathcal{C}(\tilde{\textbf{x}})}{\partial \textbf{x}_{t}^{i}}
    -\epsilon^i)\frac{\delta g(\theta, c^i)}{\delta \theta}],
\end{split}
\end{equation}
where $\{\epsilon^i\}_{i=1}^V$ are noises during score matching for different views. 
The proof can be found in the Appendix. 
To intuitively compare SDS with JSD, we sample multi-view images from the forward pass of pre-trained diffusion model~\cite{sd2022} as shown in Fig.~\ref{fig:distribution}\textcolor{red}{(c)}. 
For each view $\textbf{x}_t$, we randomly select $\textbf{x}_t^{\prime}$ from a different view and adopt the binary classification model presented in Sec.~\ref{sec:view} as the energy function to measure coherence across the two views. 
The results illustrate that JSD significantly enhances the correspondence with directional prompts and consistency across different views, particularly for the initially biased views such as side and back views. These theoretical and empirical results prove that the multi-face Janus issue in SDS is rooted in the view-agnostic and view-biased 2D image distribution, a challenge effectively mitigated by JSD.

\subsection{Universal View-Aware Models as Energy Function}\label{sec:view}
JSD requires an energy function $\mathcal{C}(\tilde{\textbf{x}},\tilde{\textbf{c}})$ to measure coherence across denoised images, as presented in Eq.~\eqref{eq:energy}. This energy function plays a crucial role in assessing the consistency between different views in image distribution. To demonstrate the compatibility of JSD, we employ three different types of models trained for various representative multi-view tasks.

\begin{figure}[t]
  \centering
   \includegraphics[width=1.0\linewidth]{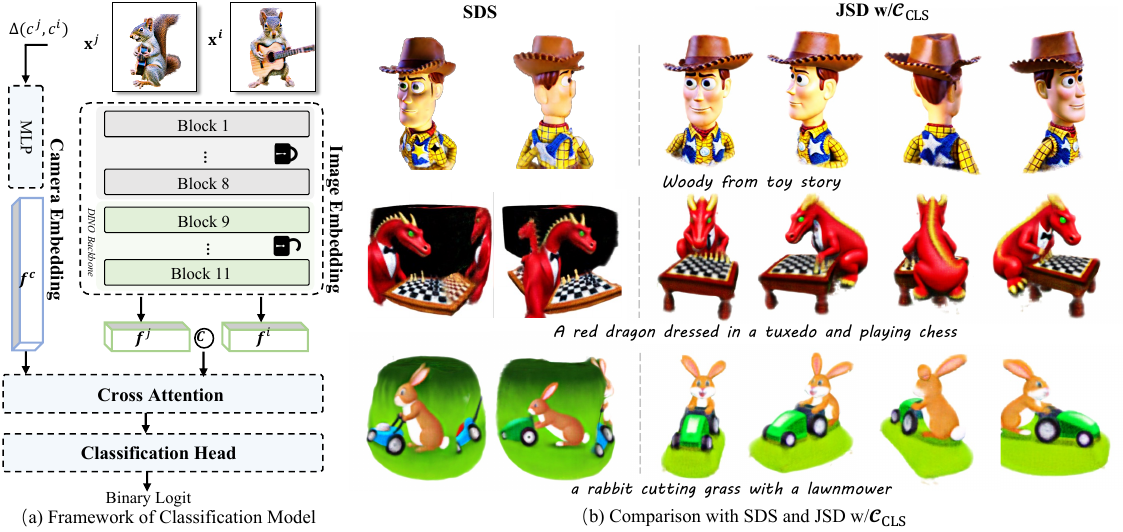}
    \vspace{-6mm}
   \caption{\textbf{Illustration of the binary classification model and qualitative results with JSD.} 
   (a) The classification model $M_{\text{CLS}}$ produces the binary logit to measure the consistency between two input views $x^i$ and $x^j$. (b) JSD integrated with the classification model effectively alleviates Janus issues compared to SDS.
   }
   \label{fig:discrimination}
\end{figure}

\begin{figure*}[ht]
  \centering
   \includegraphics[width=1\linewidth]{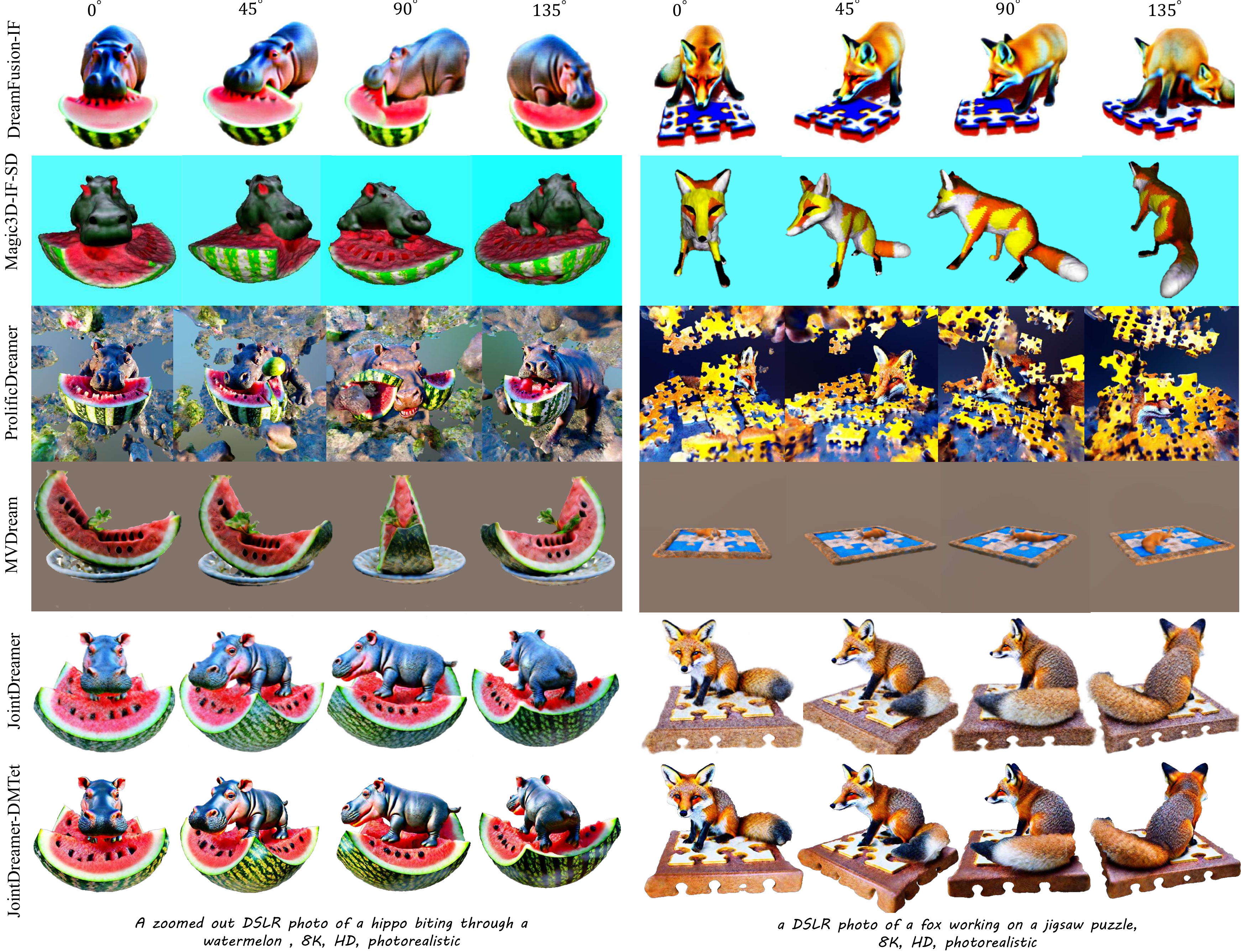}
   \vspace{-6mm}
   \caption{\textbf{Comparison of text-to-3D generation.} See Appendix for more results.}
   \label{fig:qualitycomparison}
\end{figure*}

\paragraph{Binary classification model $M_{\text{CLS}}$. }
The binary classification model $M_{\text{CLS}}$ is designed to classify the content consistency between two input images based on their relative camera pose. To ensure computational efficiency, we introduce a dedicated classification model, as shown in Fig.~\ref{fig:discrimination} \textcolor{red}{(a)}. The training process takes two days with a single A800 GPU on Objaverse dataset~\cite{deitke2023objaverse}. Further details can be found in the appendix. The classification model $M_{\text{CLS}}$ processes pairs of images $\textbf{x}^{i}$ and $\textbf{x}^{j}$ captured from different viewpoints $c^{i}$ and $c^{j}$. It extracts image features using the DINO-ViT/s16 backbone~\cite{caron2021emerging}. These image features are conditioned on the camera feature obtained through MLP layers from the relative camera transformation matrix $\Delta(c^{j},c^{i})$. Finally, the classification head produces the binary score. 
To incorporate with JSD, we only consider neighboring views in $V$ as image pairs for coherence measurement, which is denoted as:
\begin{equation}\small
    \mathcal{C}_{\text{CLS}}(\tilde{\textbf{x}},\tilde{\textbf{c}}) = \sum_{\mathclap{i,j\in 1, \dots, V; i \neq j}} M_{\text{CLS}}(\textbf{x}^{i}_t,\textbf{x}^{j}_t, \Delta(c^{j},c^{i})),
\end{equation}
where the higher logit indicates the stronger geometric consistency. 

\paragraph{Image-to-image translation model $M_{\text{I2I}}$. }
The image-to-image translation model $M_{\text{I2I}}$ is tailored for novel view synthesis~\cite{liu2023zero, long2023wonder3d, shi2023zero123++, liu2023syncdreamer}. We employ the most recent model, Wonder3D~\cite{long2023wonder3d} as $M_{\text{I2I}}$, which is a viewpoint-conditioned image translation model and generates consistent content in the target viewpoint. When integrated with JSD, a random reference view $\textbf{x}^{\textbf{ref}}$ is selected from the set of 3D rendered images. Then we input the relative camera transformation $\Delta(c^{i},c^{\textbf{ref}})$ and rendered images to $M_{\text{I2I}}$.
The measure of consistency is determined by calculating the reconstruction loss between the synthesized new image and its corresponding rendered image:
\begin{equation}\small
    \mathcal{C}_{\text{I2I}}(\tilde{\textbf{x}},\tilde{\textbf{c}}) = - \sum_{\mathclap{i\in 1, \dots, V}}|| M_{\text{I2I}}(\textbf{x}^{\textbf{ref}}_{t}, \Delta(c^{i},c^{\textbf{ref}})) - \textbf{x}_{t}^{i} ||_{2}^{2},
\end{equation}
where a smaller reconstruction loss indicates stronger geometric consistency under the estimation of $M_{\text{I2I}}$.

\paragraph{Multi-view synthesis model $M_{\text{MVS}}$. } 
The multi-view synthesis model is designed to generate multiple images conditioned on text prompts and camera poses, wherein we employ very recent work MVDream~\cite{shi2023mvdream} as $M_{\textbf{MVS}}$. We compute the reconstruction loss for multiple generative views and rendered views:

\begin{equation}\small
    \mathcal{C}_{\textbf{MVS}}(\tilde{\textbf{x}},\tilde{\textbf{c}}) = - || M_{\textbf{MVS}}(y, \tilde{\textbf{c}}) - \tilde{\textbf{x}}||_{2}^{2}
\end{equation}
The smaller reconstruction loss signifies better geometric consistency within $M_{\textbf{MVS}}$. 
These view-aware models measure the coherence across views according to their own 3D-aware insights. Incorporated with JSD, they provide distinct constraints, resulting in different 3D generations that nonetheless contribute to enhanced geometric consistency. 
We believe that JSD can be adapted to more universal view-aware models, which enables us to progressively redefine the benchmark of 3D generation with the advancement of multi-view tasks.

\begin{figure}[t]
  \centering
   \includegraphics[width=1\linewidth]{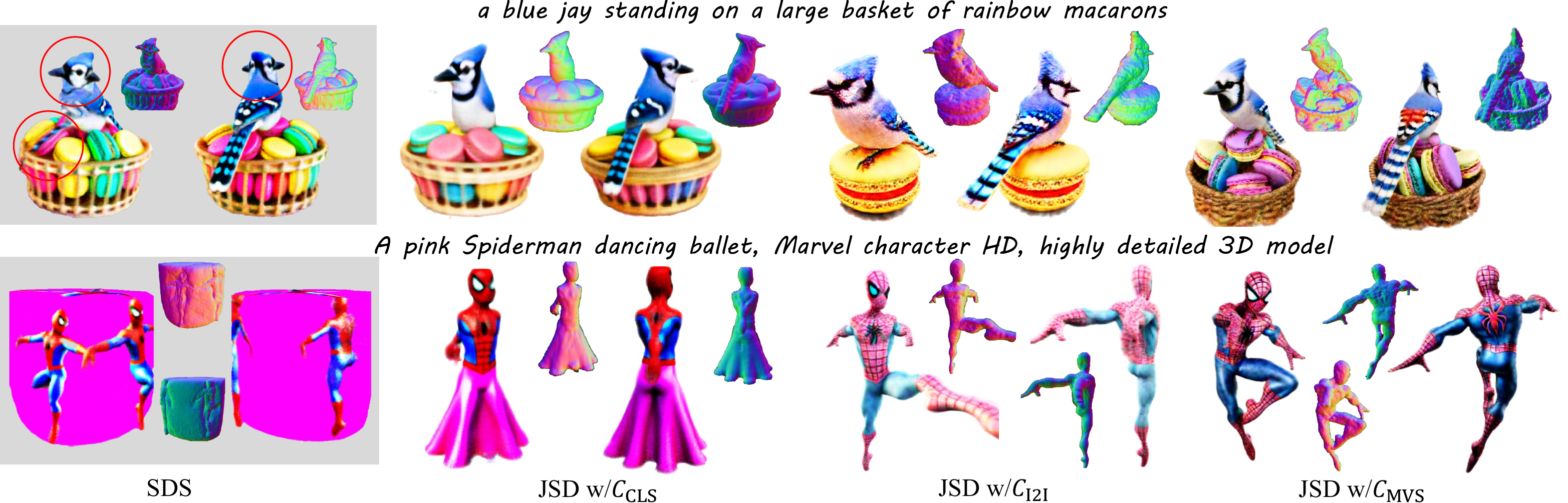}
\vspace{-4mm}
   \caption{\textbf{Ablations on JSD incorporated with different energy function $\mathcal{C}$}. CLS: Binary Classification model; I2I: Image-to-Image Translation model (Wonder3D~\cite{long2023wonder3d}); MVS: Multi-View Image Synthesis model (MVDream~\cite{shi2023mvdream}).}
   \label{fig:guidance}
\end{figure}

\subsection{Framework of JointDreamer}\label{sec:dreamer}

Building upon the JSD optimization, we propose the overall framework JointDreamer. The optimization is based on neural radiance field (NeRF)~\cite{mildenhall2021nerf}, adopting Instant-NGP~\cite{muller2022instant} with volume renderer. We adopt the multi-view synthesis model $M_{\text{MVS}}$ as the energy function to integrate with JSD in JointDreamer. During optimization training, we utilize the common techniques including time-annealing and resolution scaling-up following~\cite{wang2023prolificdreamer, shi2023mvdream}. Besides, we propose two novel techniques to further enhance the generation quality, including a \textit{Geometry Fading} scheme and a \textit{Classifier-Free Guidance Scale (CFG) Switching} strategy.

\paragraph{Geometry Fading.}
We aim to shift attention between geometric structure and texture details during optimization. Specifically, starting from iteration 5K, we reduce the learning rate of the density network of NeRF from $1e-2$ to $1e-6$ and set orientation loss to 0. Consequently, it benefits geometric convergence in the early phase of optimization, while allowing for decreased attention on geometry and increased attention on texture enhancement in the later stages.

\paragraph{CFG Switching.}
CFG scheduling strategies have been employed in 2D domain~\cite{sanghi2023clip,ho2022imagen} to enhance quality. In this work, we propose to modify the CFG scale $s$ during the training for 3D generation. 
We are motivated by the observation that a large CFG scale can lead to accelerated geometric convergence but may result in under-optimized geometry and distorted texture.
Unlike the annealed CFG approach in CLIP-Sculptor~\cite{sanghi2023clip}, our increasing strategy prioritizes texture while maintaining accurate geometry. Specifically, a smaller $s=30$ is employed in the early stages to preserve shape integrity, which allows for stronger coherence guidance from JSD. After 5K iterations, we increase the $s$ to 50, enhancing texture fidelity and overall quality.

%% file: tex/4_experiment.tex
\section{Experiment}
In this section, we present the text-to-3D generation results of JointDreamer with qualitative and quantitative evaluations, illustrating state-of-the-art performance. We also make further ablation analysis on the proposed JSD. More training and evaluation details can be found in the Appendix.

\subsection{Text-to-3D Generation}
\begin{figure}[t]
  \centering
   \includegraphics[width=1\linewidth]{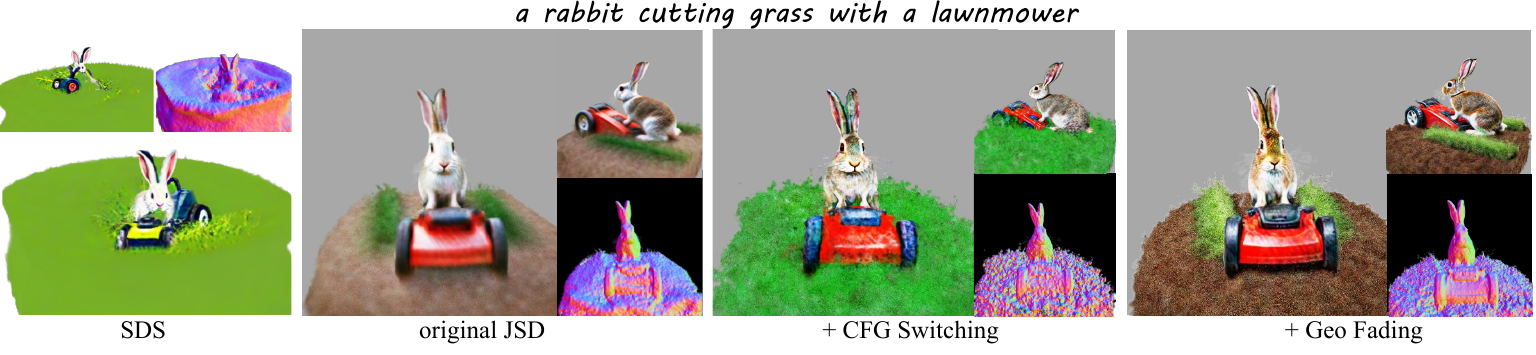}
   \vspace{-6mm}
   \caption{\textbf{Ablations on CFG Switching and Geometry Fading techniques in JointDreamer pipeline}, demonstrating the effectiveness of quality enhancement.}
   \label{fig:ablationcfg}
\end{figure}

\input{tab/quantity}

\paragraph{Qualitative Comparisons.}
We compare with several representative baselines on threestudio~\cite{threestudio2023} project, which modularizes the text-to-3D framework allowing for fair comparisons by ablating individual components. The generative samples are
shown in Fig.~\ref{fig:qualitycomparison}. 
\textit{(\romannumeral1) DreamFusion~\cite{dreamfusion22}:} Compared to DreamFusion which utilizes traditional SDS optimization, JointDreamer significantly improves the Multi-Face Janus issues in the generations of DreamFusion by introducing an energy term to ensure inter-view coherence. 
\textit{(\romannumeral2) Magic3D~\cite{magic3d22}:} Magic3D introduces a two-stage generation pipeline, transferring NeRF to DMTet in the second stage to enhance generation quality. We also transfer NeRF to DMTet as JointDreamer-DMTet, showcasing consistent superiority in geometry and texture quality by utilizing JSD optimization instead of SDS.
\textit{(\romannumeral3) ProlificDreamer~\cite{wang2023prolificdreamer}: }ProlificDreamer presents variational score distillation (VSD) as a variant of score distillation function. While VSD enhances photorealism in 3D renderings by introducing a LoRA model, the ill-posed association of pose and images during LoRA training deepens the geometric inconsistency of 3D representation, resulting in severe multi-view artifacts in generations. In contrast, JointDreamer with JSD achieves geometric consistency while maintaining high-fidelity texture quality.  \textit{(\romannumeral4) MVDream~\cite{shi2023mvdream}:} Compared to direct distillation from a finetuned model as MVDream, JointDreamer employs view-aware models as the coherence constraint in JSD, while still inherit the generalization capabilities of original diffusion models~\cite{sd2022}. The results indicate that JointDreamer mitigates the overfitting issues of MVDream, which accurately responds to complex input text and enhances the 3D consistency of generations.

\paragraph{Quantitative Comparisons. }
Following ~\cite{dreamfusion22,jain2022zero}, we evaluate CLIP Score~\cite{hessel2021clipscore}, CLIP R-Precision~\cite{park2021benchmark} and user preference on the object-centric caption subset of MS-COCO~\cite{lin2014microsoft} with 153 prompts to measure the text congruence and generative quality. For computational efficiency, we generate each 3D asset using 5K iterations of 64$\times$64 rendered images and render 20 images per caption for evaluation. CLIP ViT-B/32 is adopted as the feature extractor for Clip Score and Clip R-Precision.
As illustrated by the results in Table~\ref{tab-shape}, JointDreamer can outperform all baselines on CLIP Score and CLIP R-Precision by large margins. Specifically, JointDreamer achieves an improvement of the R-Precision by 60.8\% and 54.9\% over DreamFusion and MVDream, demonstrating its superior corresponding to textual description. Notably, the severe Janus artifacts in ProlificDreamer compromise the quality of rendering with noisy background and semantic distortion, resulting in the lowest R-Precision. 
We also conduct a user study about shape preference on these prompts in Table~\ref{tab-shape}.

\subsection{Ablation Analyses}
\paragraph{Ablations on Energy Functions.}
Sec.~\ref{sec:view} discusses the varying inter-view
\input{tab/quantity_2}
coherency measurements provided by view-aware models trained.
When incorporated with JSD, these models have distinct impacts on 3D generations, as shown in Fig.~\ref{fig:guidance}. The binary classifier effectively corrects inaccurate geometric structures in SDS. However, it cannot introduce additional imaginative elements as a discriminative model, resulting in oversaturated and monotonous textures. In contrast, as generative models, Wonder3D~\cite{long2023wonder3d} and MVDream~\cite{shi2023mvdream} employ reconstruction loss to estimate 3D consistency. Hence, they not only guide geometric structural modifications but also influence texture quality.
we further conduct a quantitative comparison using 16 complex multi-Janus prompts, as outlined in Table~\ref{tab-janusrate}. The experimental setup details can be found in the Appendix. The results indicate that our JSD consistently mitigates Janus artifacts across different view-aware models, with only a slight increase in computational requirements compared to SDS. We find that the energy function $\mathcal{C}_{\text{I2I}}$ derived from the image-to-image translation model exhibits poor performance, likely attributed to a mismatch in camera range and inaccurate translated results.

\paragraph{Analysis on Geometry Fading and CFG Switching Mechanisms.}
We conduct incremental ablations on our proposed 
techniques in JointDreamer, including the Geometry Fading scheme and  CFG Switching strategy. As shown in Fig.~\ref{fig:ablationcfg}, increasing the CFG value enhances texture detail compared to the original JSD, but over-optimizes the geometry, resulting in a bumpier shape. Compared to ``+CFG Switching'', the Geometry fading effectively protects shape when larger CFG guidance.
We also conduct a quantitative evaluation on a 30\% MS-COCO subset in Table~\ref{tab-abs}. JSD demonstrates superior texture quality compared to SDS, as evidenced by the Clip Score (CS) and FID metrics. The two proposed mechanisms further enhance the texture quality.

\paragraph{Discussions on Training Loss.} To make further comparisons with JSD
\begin{wrapfigure}{r}{0.5\textwidth}
\vspace{-8mm}
  \centering
   \includegraphics[width=1\linewidth]{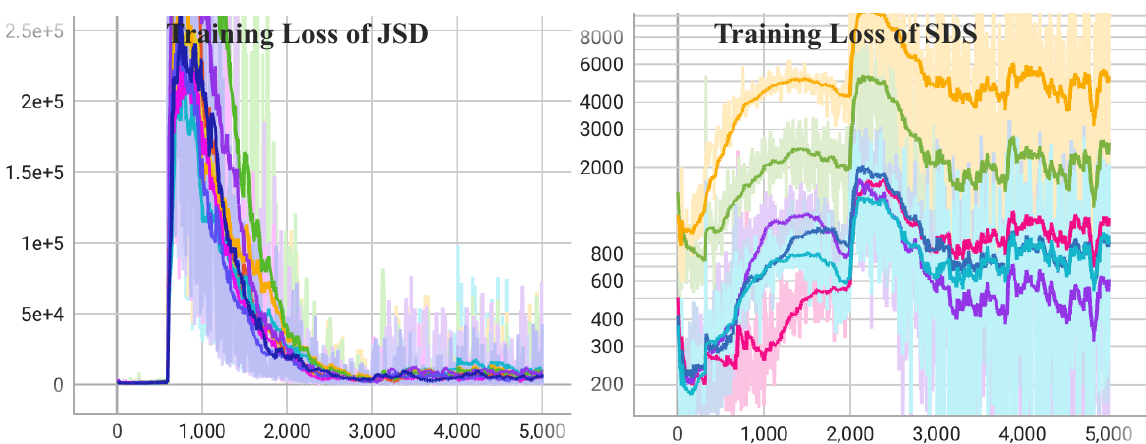}
   \vspace{-5mm}
   \caption{\textbf{Training loss comparisons of JSD and SDS.} JSD eliminates randomness fluctuations in SDS convergence.}
   \label{fig:sdloss}
      \vspace{-5mm}
\end{wrapfigure}
and SDS, we aggregate training losses from multiple prompts on two optimization functions and visualize the training loss curve as illustrated in Fig.~\ref{fig:sdloss}. We observe that SDS experiences significant fluctuations due to the randomness of single-view optimization. In contrast, JSD can converge gradually and smoothly, demonstrating the introduction of multi-view optimization with inter-view coherence in JSD can reduce the randomness of optimization and contribute to better convergence for 3D representation.

\begin{figure*}[t]
  \centering
   \includegraphics[width=1.\linewidth]{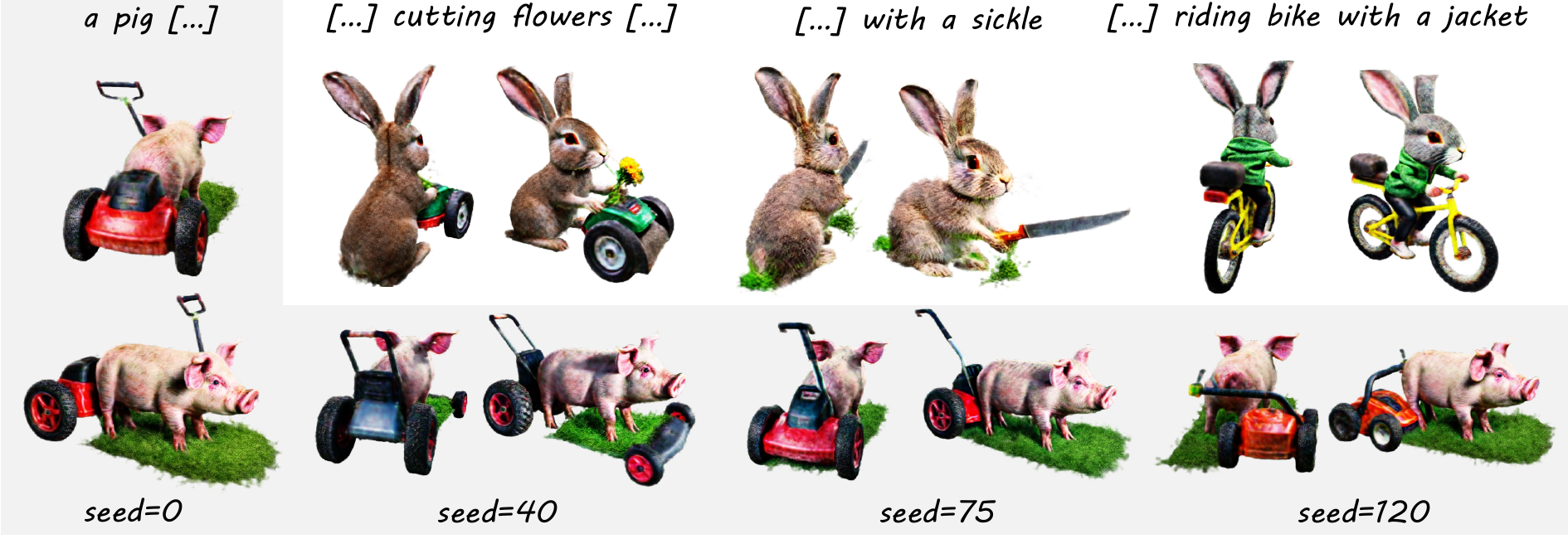}
   \vspace{-6mm}
   \caption{\textbf{Discuss the impact of prompts and random seeds,} demonstrating the robustness of JointDreamer. \textit{[...]} means the same content as the prompt in Fig.~\ref{fig:ablationcfg}.
   }
   \vspace{-1mm}
   \label{fig:robust}
\end{figure*}

\paragraph{Discussions on Robustness for Prompts and Random Seed.} To show the impact of seed and prompt, 
we modify seeds and key prompt components in Fig.~\ref{fig:ablationcfg}, such as subject, object and verb.
Results in Fig.~\ref{fig:robust} demonstrate the robustness of JointDreamer for different seed. Note that the default seed is 0.

\paragraph{The Effectiveness of Classification Model.} 
The classification model surpasses MVDream in training speed by a factor of 48. Fig.\ref{fig:discrimination} \textcolor{red}{(b)} provides additional high-quality results obtained using JSD in conjunction with the binary classification model $M_{\text{CLS}}$. Furthermore, we conduct an ablation study by replacing $\frac{\partial \mathcal{C}(\tilde{\textbf{x}})}{\partial \textbf{x}_{t}^{i}}$ in Eq.\ref{eq:JSD} with a randomly generated value between $[0,1]$. However, it yields shapeless results due to unrelated disturbances in the optimization process. These findings highlight the significance of the proposed classification model in achieving a balance between computational efficiency and generation quality.

\begin{figure}[t]
  \centering
   \includegraphics[width=0.98\linewidth]{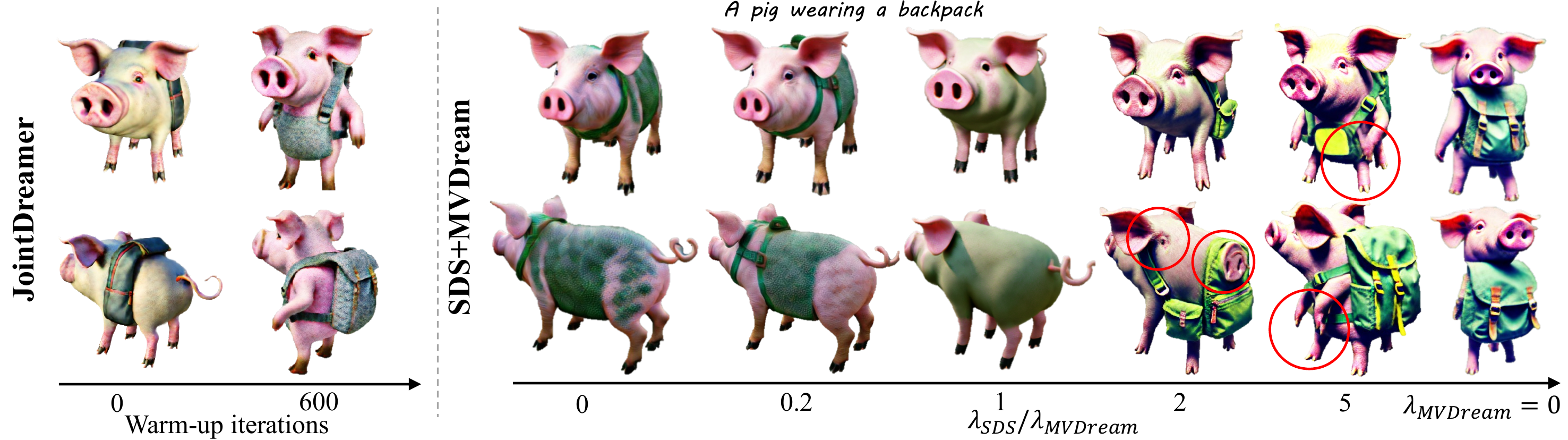}
   \vspace{-2mm}
   \caption{\textbf{Comparison of JointDreamer and SDS+MVDream}, illustrating the superiority of JointDreamer in mitigating the Janus problem. SDS+MVDream consistently exhibits semantic missing or Janus issues with different weight options. }
   \label{fig:comp2d3d}
   \vspace{-2mm}
\end{figure}

\paragraph{More Comparison with MVDream.}
View-aware models can be incorporated into 3D generation by combining them with JSD or using them directly with SDS. In this discussion, we use MVDream~\cite{shi2023mvdream} as an example to demonstrate the superiority of JSD.
MVDream can generate consistent shapes when calculating the SDS loss directly. However, it may miss components in the complete input text due to its fine-tuning on limited 3D data, as illustrated in Fig.~\ref{fig:qualitycomparison}. To address this limitation, a straightforward approach is to ``SDS+MVDream''. 

However, balancing the impact of SDS and MVDream is challenging. As shown in Fig.~\ref{fig:comp2d3d}, setting $\lambda_{\textbf{SDS}}$ to 0 degrades to MVDream alone, while setting $\lambda_{\textbf{MVDream}}$ to 0 yields a combination similar to DreamFusion~\cite{dreamfusion22}. Achieving a balanced impact between SDS and MVDream through a simple combination is difficult. When $\lambda_{\textbf{MVDream}}$ is large, textual consistency remains constrained, whereas decreasing $\lambda_{\textbf{MVDream}}$ leads to the Janus problem. This is due to gradient misalignment across multiple views in the ``SDS+MVDream'' combination, as SDS lacks multi-view information and cannot derive our objective outlined in Eq.~\ref{eq:JSD}.
In contrast, JSD based on the joint image distribution provides supervision for text consistency and high-fidelity texture. JSD also promotes inter-view consistency by introducing an energy term with the view-aware model. 

%% file: tab/quantity.tex
\begin{minipage}{0.58\textwidth}
\vspace{-3mm}
\captionof{table}{\textbf{Quantitative results on texual consistency and user preference}, tested on object-centric subset of MS-COCO~\cite{lin2014microsoft}.}
\resizebox{\linewidth}{!}{
\begin{centering}
\begin{tabular}{c|ccc}
\toprule
Method & CLIP Score $\uparrow $ & R-Precision $\uparrow $& User Study$\uparrow $  \tabularnewline\midrule 
DreamFusion~\cite{dreamfusion22}  & 20.1 &27.7& 18.2\tabularnewline
ProlificDreamer~\cite{wang2023prolificdreamer}& 25.0 & 18.7& 16.2  \tabularnewline
MVDream~\cite{shi2023mvdream}& 20.8 & 33.6  & 23.5 \tabularnewline
JointDreamer& \textbf{27.7} & \textbf{88.5} & \textbf{42.1}  \tabularnewline\bottomrule
\end{tabular}
\end{centering}
}
\label{tab-shape}
\end{minipage}
\begin{minipage}{0.4\textwidth}
\vspace{-3mm}
\captionof{table}{\textbf{Ablation study on CFG Switching (CFGS) and Geometry Fading (GF)}.}
\resizebox{\linewidth}{!}{
\begin{centering}
\label{tab:quantitative1}
\begin{tabular}{cccc|cc}
\toprule
SDS & JSD & CFGS & GF & CLIP Score$\uparrow$ & FID$\downarrow$ \tabularnewline\midrule 
$\checkmark$ &  &  &  & 20.0&429.2\tabularnewline
 & $\checkmark$ &  &  & 27.6 & 360.7\tabularnewline
  & $\checkmark$ & $\checkmark$ & & 28.2 & 357.6\tabularnewline
   & $\checkmark$ & $\checkmark$ &  $\checkmark$& 28.8 & 353.9\tabularnewline\bottomrule

\end{tabular}
\end{centering}
}
\label{tab-abs}
\end{minipage}

%% file: tab/quantity_2.tex
\begin{wraptable}{r}{0.45\textwidth}
\vspace{-8mm}
\caption{\textbf{Quantitative comparison of energy functions,} showcasing the effectiveness of JSD in mitigating Janus artifacts with comparable computational efficiency to SDS.}
\resizebox{\linewidth}{!}{
\begin{centering}
\begin{tabular}{c|ccc} %
\toprule
Methods & Janus Rate $\downarrow$ & GPU Memory $\downarrow$ & Train Time $\downarrow$\tabularnewline\midrule 
SDS & 100\% & 16.1 G  & 50 min.\tabularnewline
JSD w/$\mathcal{C}_{\textbf{CLS}}$ & 12.5\% & 22.1 G & 80 min.  \tabularnewline
JSD w/$\mathcal{C}_{\textbf{I2I}}$ & 31.2\% & 16.0 G & 119 min. \tabularnewline
JSD w/$\mathcal{C}_{\textbf{MVS}}$ &  6.2\% & 19.4 G & 54 min. \tabularnewline\bottomrule

\end{tabular}
\end{centering}
}
\label{tab-janusrate}
\vspace{-6mm}
\end{wraptable}

%% file: tex/5_conclusion.tex
\section{Conclusion}
In this work, we introduce Joint Score Distillation (JSD) as a new paradigm for text-to-3D generation, which conducts multi-view optimization jointly and accounts for inter-view coherence. We demonstrate that JSD can significantly enhance 3D coherence while maintaining generalizability. With other proposed techniques, our overall framework, JointDreamer, is capable of geometric-consistent and high-fidelity 3D generation adhering to complex text input. 

\noindent\textbf{Limitations.}
While the training time of JointDreamer is comparable to existing SDS pipelines, there is room for improvement in terms of acceleration. Future work will explore alternative 3D representations, such as 3D Gaussian~\cite{kerbl20233d}. Additionally, JSD utilizes view-aware models to ensure geometry consistency and mitigate the impact of limited data. However, view-aware models still require 3D data for training, and further efficient 3D data collection or reconstruction from multi-view images is also worth investigating.

\noindent\textbf{Acknowledgement.}
This research has been made possible by funding support provided to Dit-Yan Yeung by the Research Grants Council of Hong Kong under the Research Impact Fund project R6003-21.

%% file: sub_tex/supp.tex
\clearpage
\setcounter{figure}{0}
\renewcommand{\thefigure}{A\arabic{figure}}
\appendix

This supplementary material consists of five parts, including technical details of the experimental setup (Sec.~\ref{sec-details}), the derivation of Joint Score Distillation (JSD) (Sec.~\ref{sec-jsd}),  additional ablation analysis (Sec.~\ref{sec-abla}), additional experimental results (Sec.~\ref{sec-results}) and the Janus prompt list (Sec.~\ref{sec-prompt}).

\section{Experimental Setup}\label{sec-details}

\subsection{Details of JointDreamer Pipeline.}
In our main text, we adopt MVDream $\mathcal{M}_{\text{MVS}}$ as the energy function for the overall JointDreamer pipeline. Since MVDream fine-tunes on SD-V2.1, we retain SD-V2.1 as a diffusion model.
The whole training procedure includes 6k iterations, taking around 1.5 h with batch size 4 on 1 Nvidia Tesla A800 GPU.
Specifically, we warm up NeRF for the initial 600 training iterations with SDS and adopt JSD for the remaining iterations. 
We adopt the common time-annealing and resolution-increasing tricks from the open-source implementation, together with the two proposed mechanisms including the Geometry Fading scheme and Classifier-Free Guidance (CFG) Scale switching strategy. 
We set $t=0.98$ with resolution 64 for the first 3k iterations and then anneal into $t \sim U(0.02, 0.50)$ with resolution 256 for the extra 2k iterations. Starting from iteration 5k, we scale up the resolution to 512 and conduct the two proposed mechanisms, where the learning rate of the density network is reduced from $1e-2$ to $1e-6$ and the CFG scale is switched from 30 to 50. The Geometry Fading scheme and Classifier-Free Guidance (CFG) Scale switching strategy allow greater influence from coherence guidance in JSD on geometry optimization in the early training stages and enhance the fidelity of textures in later stages. 

\subsection{Details of Binary Classification Model.}

In this part, we will elaborate on the model architecture and training procedure of the binary classification model that is discussed in Sec.4.2 in the main paper.

\paragraph{Model Architecture.}
We build the model based on the DINO framework. 
Specifically, we employ ViT-s16 as the backbone for extracting image features. The backbone is initially pre-trained following the DINO method, and during training, the first 9 blocks of the backbone are frozen.
Besides, we use a 4-layer MLP with 256 hidden layer channels to extract the relative camera embedding of the transformation matrix between input images, which captures the camera-specific information.
Next, we calculate the cross-attention between camera embedding and the concatenated image features of input image pairs. This cross-attention mechanism generates a residual feature input, combined with the concatenated image features as the final feature.
Finally, the combined features are fed into the classification head consisting of a 3-layer MLP, which produces the classification logit prediction for input image pairs.

\paragraph{Training Procedure.}
For training data, we use rendered images from\input{sub_fig/wrap_clstrain}Objaverse~\cite{deitke2023objaverse} following Zero-1-to-3~\cite{liu2023zero}. For the binary classification training objective, we adopt the pairs of images from the same object equipped with the correct camera pose as the positive samples and assign the image pairs from different objects or incorrect relative camera poses as negative samples. Before training, we prepare the index list of positive and negative pairs for efficient training. During training, we randomly sample 1 million positive pairs and 1 million negative pairs from the index list as training sets. The design of the training set ensures that the classification model can identify the 3D consistency between rendered images conditioned on relative camera pose.
We adopt adamW optimizer with $5e-4$ learning rate and 0.04 weight decay. We also adopt random color jitter, gaussian blur, and polarization following DINO as data augmentation.
We use an image size of $224\times224$ and a total batch size of $640$ and train the model for 10 epochs.
The training takes about 1 day on 2 Nvidia Tesla A800 GPUs. To validate the classification accuracy, We random sample 5000 pairs as the validation set. The training loss and validation accuracy curve can be found in Fig.~\ref{fig:clstrain}. 


\subsection{Details of Text-to-3D Generation Comparison}
\paragraph{Baseline Setup. }
We implement the experiments in an open-source threestudio project and reproduce DreamFuion-IF, Magic3D-IF-SD, and ProlificDreamer as baselines following the comparisons in the main paper of MVDream. Our MVDream baseline is reproduced by its officially released code. We adopt DeepFloyd-IF~\cite{deepfloyd} as the 2D diffusion model for baseline DreamFuion-IF and the first stage of Magic3D-IF-SD following MVDream. To make a fair comparison with our JointDreamer, we equip the same batch size, resolution, and time annealing strategy with JointDreamer for DreamFuion-IF. 

\paragraph{Evaluation Details.}
We conducted a user study from 100 users on the 153 generated models from the object-centric MS-COCO subset. Each user is given 4 rendered videos with their corresponding text input from generations of different methods. We ask the users to select a preferred 3D model from four options, and then calculate the mean proportion of each method selected over all 153 prompts as the score. The higher score indicates the greater user preference. For the Clip Score and Clip R-Precision, we adopt the CLIP ViT-B/32
as the feature extractor.

\begin{figure}[t]
  \centering
   \includegraphics[width=1\linewidth]{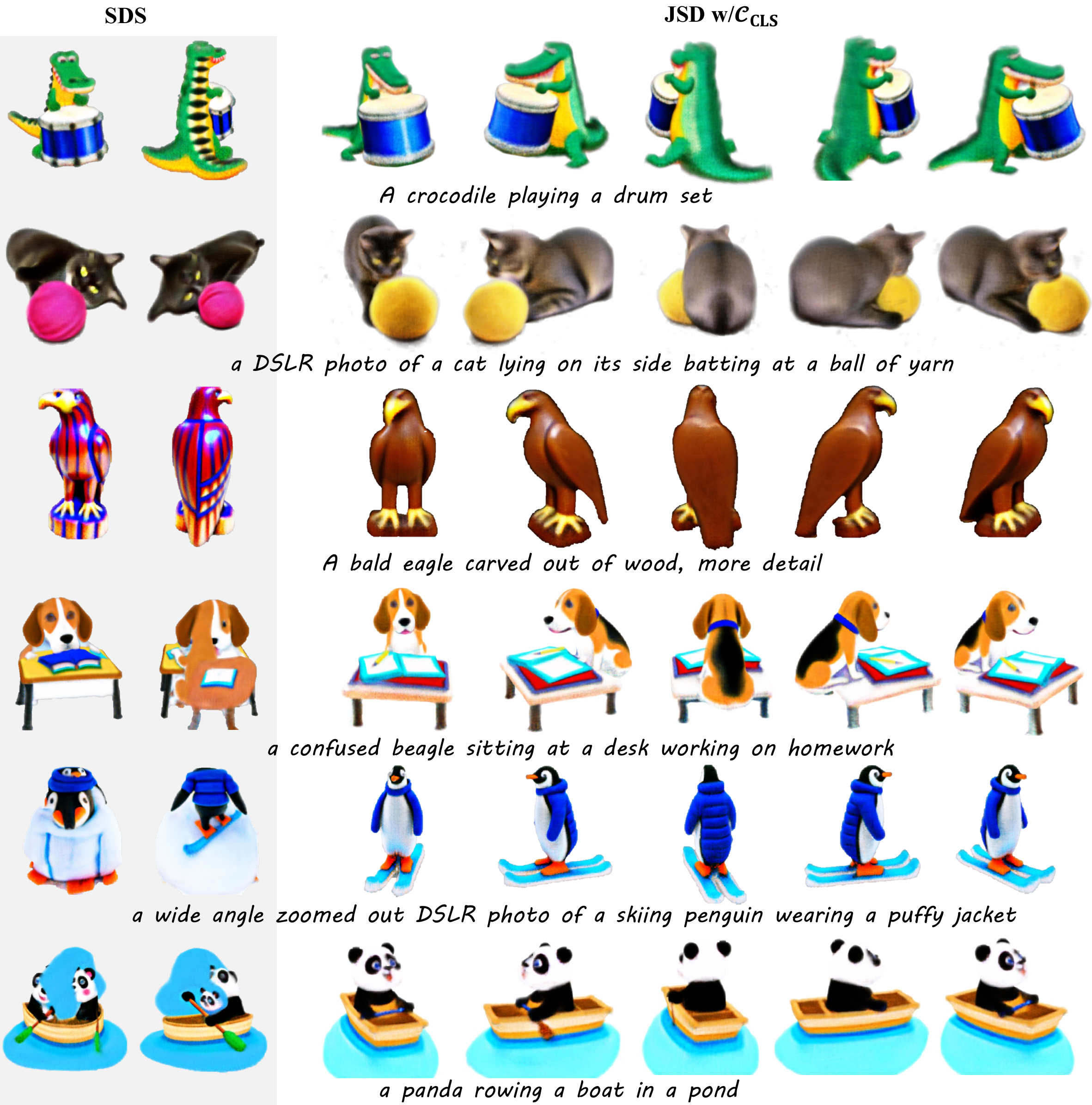}
   \vspace{-4mm}
   \caption{\textbf{More quality results of JSD with Classification Model.}}
   \label{fig:clsresults}
\end{figure}

\subsection{Details of Computational Resource Comparison}
We analyze the geometry consistency and computation efficiency of various view-aware models in main paper Table~\ref{tab-janusrate}, using 16 complex multi-Janus prompts in Sec.~\ref{sec-prompt} from the DreamFusion~\cite{dreamfusion22} library.
We maintain consistent experimental parameters, including a batch size of 4, training 5k iterations and a resolution of 64, as well as the same optimizer and time annealing hyperparameters. The only variation is in the camera parameters, which align with each view-aware model's settings. For the baseline SDS model, we adopt the DreamFusion camera parameters. We present some examples showcasing these results incorporating $\mathcal{C}_\textbf{CLS}$ in Figure~\ref{fig:clsresults}. And the results incorporating $\mathcal{C}_\textbf{MVS}$ can be found in Section~\ref{sec-results}.

\section{Theory of Joint Score Distillation}\label{sec-jsd}

Given a well-trained text-to-image diffusion model, like Stable Diffusion, the objective is to distill its knowledge into a 3D representation network parameterized by $\theta$, such as NeRF and ensures coherent 3D generations. To achieve this, we aim to model the joint rendering distribution across multiple views of $\theta$.

For ease of notation, we define $\tilde{\bm x}$ as the joint random variable comprising $\bm x^1, \ldots, \bm x^V$, which are rendered images sampled from the 3D representation $\theta$. It is important to note that these views are not independent. In a 3D model, the views are inherently connected as they originate from the same underlying 3D object. This means that the rendered images, $\bm x^1, \ldots, \bm x^V$, exhibit dependencies and correlation.

Denote the joint rendering distribution of $\tilde{\bm x}$ as $\tilde{q}^\theta$. We can still define the marginal distributions as 
\[
q^\theta(\bm x^i) = \int \tilde{q}^\theta(\tilde{\bm x})d\tilde{\bm x}^{-i},
\]
where $\tilde{\bm x}^{-i} = {\bm x^1, \ldots,\bm x^{i-1},\bm x^{i+1}, \ldots, \bm x^V}$. This marginal distribution is the same as if only a single view is considered, i.e., $V=1$. 

We can further define the log density ratio as
\[
    R(\tilde{\bm x}) = \log\frac{\tilde{q}^\theta(\tilde{\bm{x}})}{\prod_{i=1}^V q^\theta(\bm x^i) }
\]
to capture the inter-relationship among different views. Equivalently, we can write
\[
    \tilde{q}^\theta(\tilde{\bm{x}}) = \exp(R(\tilde{\bm x}))\prod_{i=1}^V q^\theta(\bm x^i). 
\]
To get the evaluations of $\tilde{\bm x}$ from the 2D diffusion model, we have 
\[
    \tilde{p}(\tilde{\bm x}) \propto \exp(\mathcal{C}(\tilde{\bm x})) \prod_{i=1}^V p(\bm x^i)
\]
since the diffusion model only takes a single image as input and different views are weighted by the introduced joint energy function $\mathcal{C}$.

Now we consider learning $\tilde{q}^\theta(\tilde{\bm x})$ such that the following Integral Kullback–Leibler (IKL) divergence is minimized along the forward diffusion process $\bm x_t = \alpha_t \bm x_0 + \sigma_t \epsilon$ where $\epsilon$ follows standard Gaussian distribution. 
\begin{align*}\small
    \min_\theta D_{\mathrm{IKL}}(\tilde{q}^\theta(\tilde{\bm x})||\tilde{p}(\tilde{\bm x})) &= \min_\theta \int_0^T w(t)\frac{\sigma_t}{\alpha_t}D_{\mathrm{KL}}(\tilde{q}_t^\theta(\tilde{\bm x})||\tilde{p}_t(\tilde{\bm x})) dt\\
    &= \min_\theta \int_0^T w(t)\frac{\sigma_t}{\alpha_t}\mathbb{E}_{\tilde{\bm x}_t \sim \tilde{q}_t^\theta}\left(\log\frac{\tilde{q}_t^\theta(\tilde{\bm x}_t)}{\tilde{p}_t(\tilde{\bm x}_t)}\right) dt.
\end{align*}
Taking gradient with respect to $\theta$ gives
\begin{align*}
    &\frac{\partial}{\partial \theta}  D_{\mathrm{IKL}}(\tilde{q}^\theta(\bm x)||\tilde{p}(\bm x))\\
    &= \int_0^T w(t)\frac{\sigma_t}{\alpha_t}\frac{\partial}{\partial \theta} \mathbb{E}_{\tilde{\bm x}_t \sim \tilde{q}_t^\theta}\left(\log\frac{\tilde{q}_t^\theta(\tilde{\bm x}_t)}{\tilde{p}_t(\tilde{\bm x}_t)}\right) dt\\
    &= \int_0^T w(t)\frac{\sigma_t}{\alpha_t} \mathbb{E}_{\tilde{\bm x}_t \sim \tilde{q}_t^\theta} \left[\frac{\partial}{\partial \tilde{\bm x}_t}\left(\log\frac{\tilde{q}_t^\theta(\tilde{\bm x}_t)}{\tilde{p}_t(\tilde{\bm x}_t)}\right) \frac{\partial \tilde{\bm x}_t}{\partial \theta} + \frac{\partial}{\partial \theta}\log \tilde{q}_t^\theta(\bm x)|_{\bm x =\tilde{\bm x}_t} \right]dt \\
    &:= A+B.\\
\end{align*}
The term $B$ vanishes since
\begin{align*}
    B &= \int_0^T w(t)\frac{\sigma_t}{\alpha_t} \mathbb{E}_{\tilde{\bm x}_t \sim \tilde{q}_t^\theta}  \frac{\partial}{\partial \theta}\log \tilde{q}_t^\theta(\bm x)|_{\bm x =\tilde{\bm x}_t}dt\\
    &= \int_0^T w(t)\frac{\sigma_t}{\alpha_t} \mathbb{E}_{\tilde{\bm x}_t \sim \tilde{q}_t^\theta}  \frac{\frac{\partial}{\partial \theta} \tilde{q}_t^\theta(\bm x)|_{\bm x =\tilde{\bm x}_t}}{\tilde{q}_t^\theta(\tilde{\bm x}_t)}dt\\
    &= \int_0^T w(t)\frac{\sigma_t}{\alpha_t} \int  {\frac{\partial}{\partial \theta} \tilde{q}_t^\theta(\bm x)|_{\bm x =\tilde{\bm x}_t}}dt\\
    & = \int_0^T w(t)\frac{\sigma_t}{\alpha_t} \frac{\partial}{\partial \theta} \int { \tilde{q}_t^\theta(\bm x)}dt\\
    & = 0
\end{align*}
The term $A$ is the score distillation loss 
\begin{align*}
    A =  \int_0^T w(t)\frac{\sigma_t}{\alpha_t} \mathbb{E}_{\tilde{\bm x}_0 \sim \tilde{q}_0^\theta, \tilde{\epsilon}}\left(\nabla\log{\tilde{q}_t^\theta(\tilde{\bm x}_t)} - \nabla\log{\tilde{p}_t(\tilde{\bm x}_t)}\right) \frac{\partial \tilde{\bm x}_t}{\partial \theta} dt, 
\end{align*}
where $\tilde{\epsilon} = (\epsilon^1, \ldots, \epsilon^V)$ are the noises along the forward diffusion process.  
Putting things together we have
\begin{align*}
    &\frac{\partial}{\partial \theta}  D_{\mathrm{IKL}}(\tilde{q}^\theta(\bm x)||\tilde{p}(\bm x)) =  \mathbb{E}_{\tilde{\bm x}_0 \sim \tilde{q}_0^\theta, \tilde{\epsilon}, t} \left[w(t) \frac{\sigma_t}{\alpha_t} \left(\nabla\log{\tilde{q}_t^\theta(\tilde{\bm x}_t)} - \nabla\log{\tilde{p}_t(\tilde{\bm x}_t)}\right) \frac{\partial \tilde{\bm x}_t}{\partial \theta}\right]
\end{align*}
Notice that the NeRF rendering is a deterministic process given the view information. Therefore, the conditional distribution and marginal distribution coincide, i.e., 
\[
    \tilde{q}_t^\theta(\tilde{\bm x}_t)\sim N(\alpha_t\tilde{\bm x}_0,\sigma_t^2), \quad \nabla\log{\tilde{q}_t^\theta(\tilde{\bm x}_t)} = -\tilde{\epsilon}/\sigma_t.
\]
On the other hand, direct score matching tells us that 
\[
    \nabla\log{{p}_t({\bm x}^i_t)} = \frac{\partial \mathcal{C}(\tilde{\bm x})}{\partial {\bm x^i_t}} - \hat{\epsilon}_{\Phi}({\bm x}^i_t, t)/\sigma_t.
\]
Finally, combining $\frac{\partial \bm{x}_t^i}{\partial\theta}=\alpha_t \frac{\partial \bm{x}_0^i}{\partial\theta}$, we have
\begin{align}\label{eq:ikl_grad}
    &\frac{\partial}{\partial \theta}  D_{\mathrm{IKL}}(\tilde{q}^\theta(\bm x)||\tilde{p}(\bm x)) =  \mathbb{E}_{\tilde{\bm x}_0 \sim \tilde{q}_0^\theta, \tilde{\epsilon}, t} \left[w(t)  \sum_{i=1}^V\left(\hat{\epsilon}_{\Phi}({\bm x}^i_t, t) - \frac{\partial \mathcal{C}(\tilde{\bm x})}{\partial {\bm x^i_t}} - \epsilon^i  \right) \frac{\partial {\bm x}_0^i}{\partial \theta}\right].
\end{align}
Now we have finished extending SDS to multiple views. As it turns out, the joint energy term  $R(\tilde{\bm x})$ does not show up in the gradient formula.

\section{Additional Ablation Study}\label{sec-abla}

\begin{figure}[t]
  \centering
   \includegraphics[width=1\linewidth]{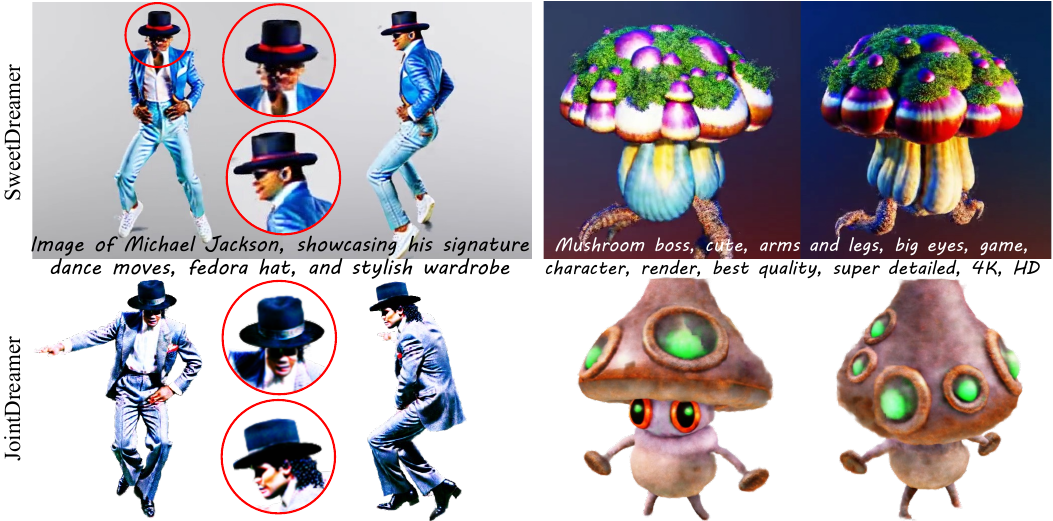}
   \vspace{-4mm}
   \caption{\textbf{Comparison with SweetDreamer.} SweetDreamer suffers from multi-faces (left) and missing components such as "legs" and "eyes" (right).}
   \label{fig:sweetdreamer}
\end{figure}

\subsection{Comparison with SweetDreamer}
We also conduct a comparison with SweetDreamer~\cite{li2023sweetdreamer}. SweetDreamer aligns geometric priors (AGP) in a finetuned diffusion model and combines AGP with SDS to address the Janus issue. In contrast, JSD improves the optimization objective of SDS with various energy functions, and AGP can be one of them. For 3D generation, Fig.~\ref{fig:comp2d3d} shows that a simple combination, like SweetDreamer's, uses more memory and complicates balancing components. Compared to SweetDreamer's demos from its website, our JointDreamer achieves better shape and text congruence without multi-faces and missing components ("arms", "big eyes") in Fig.~\ref{fig:sweetdreamer}.

\begin{figure}[ht]
  \centering
   \includegraphics[width=1\linewidth]{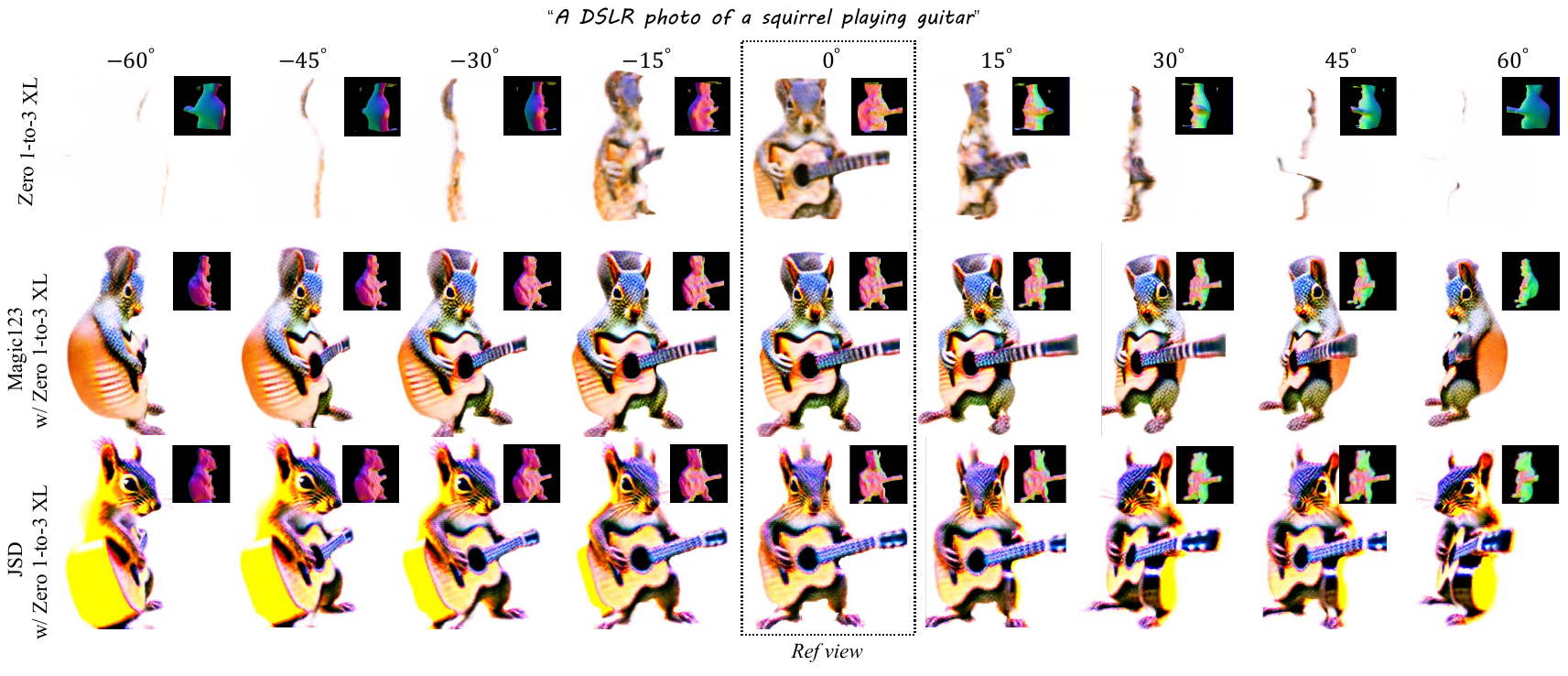}
   \vspace{-4mm}

   \caption{\textbf{Comparison with Image-to-3D methods.} Compared with two alternative methods, all employing the Zero-1-to-3 XL model, our proposed JSD  exhibits superior generative quality in novel view synthesis as evidenced by its geometric consistency.}
   \label{fig:imageto3d}
\end{figure}

\subsection{Discussions on Image-to-3D Methods}
Since the view-aware models can engage in 3D generation through SDS besides JSD, we make comparisons to showcase the superiority of JSD. 
Section 5.2 details the comparative use of MVDream, and herein, we extend this comparison to different applications of the image-to-image translation model, Zero-1-to-3 XL, which excels in image-to-3D tasks. Unlike text-to-3D approaches that generate 3D models from textual descriptions, the image-to-3D method uses a reference image to
fix the reference view and generate the remaining views. As shown in Fig.~\ref{fig:imageto3d}, we input a reference image, exemplified by the front-view rendered image of the case of  ``A DSLR photo of a squirrel playing guitar'' in Fig.~\ref{fig:quality_append1} and compare with two alternative utilizations of Zero-1-to-3 XL.  \textit{(i)Zero-1-to-3 XL~\cite{liu2023zero}}, which directly utilizes Zero-1-to-3 XL to calculate SDS loss for novel rendered views according to reference view. The overfitting generalizability of Zero-1-to-3 XL reduces the generative quality, especially for the views distant from the reference view. \textit{(ii)Magic123~\cite{qian2023magic123}}, which merges the SDS loss of SD-V2.1 and Zero-1-to-3 XL as objective function. By combining the generalizability from the original diffusion model, it can eliminate the distortion in novel views, but the effect is not satisfactory. By contrast, our JSD achieves better generation quality in novel views, where the overall geometric structure is more reasonable. Notably, when applying JSD in image-to-3D generation, we calculate the inter-view coherence between the reference view and random novel views to fix the reference view, differing from the two random novel views used in text-to-3D generation. The comparisons further illustrate that JSD provides the optimal solution to combine generalizability from 2D models and geometric understanding from 3D-aware models.

\subsection{Discussion on Failure Cases}
\begin{figure}[ht]
  \centering
   \includegraphics[width=1\linewidth]{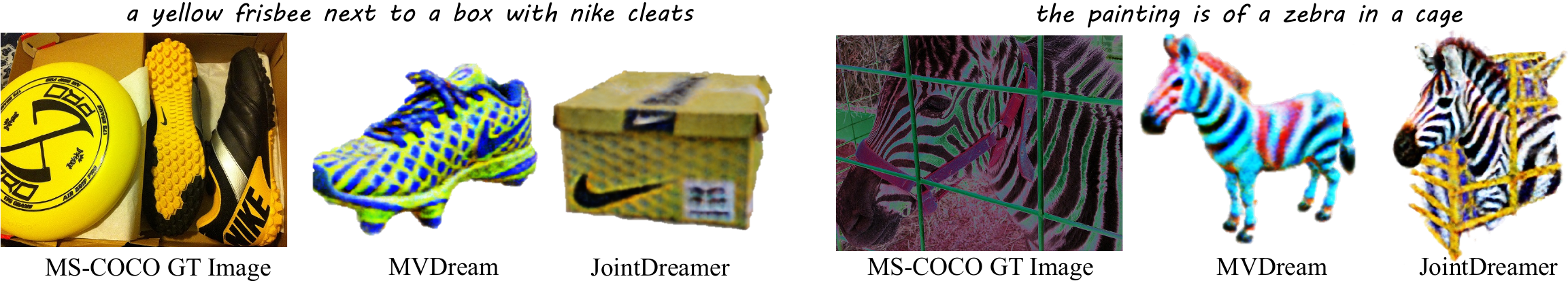}
  \vspace{-4mm}
  \caption{Failure Cases on MS-COCO Subset.}\label{fig:fail}
   \label{fig:sweetdreamer}
\end{figure}

\begin{figure}[ht]
  \centering
   \includegraphics[width=1\linewidth]{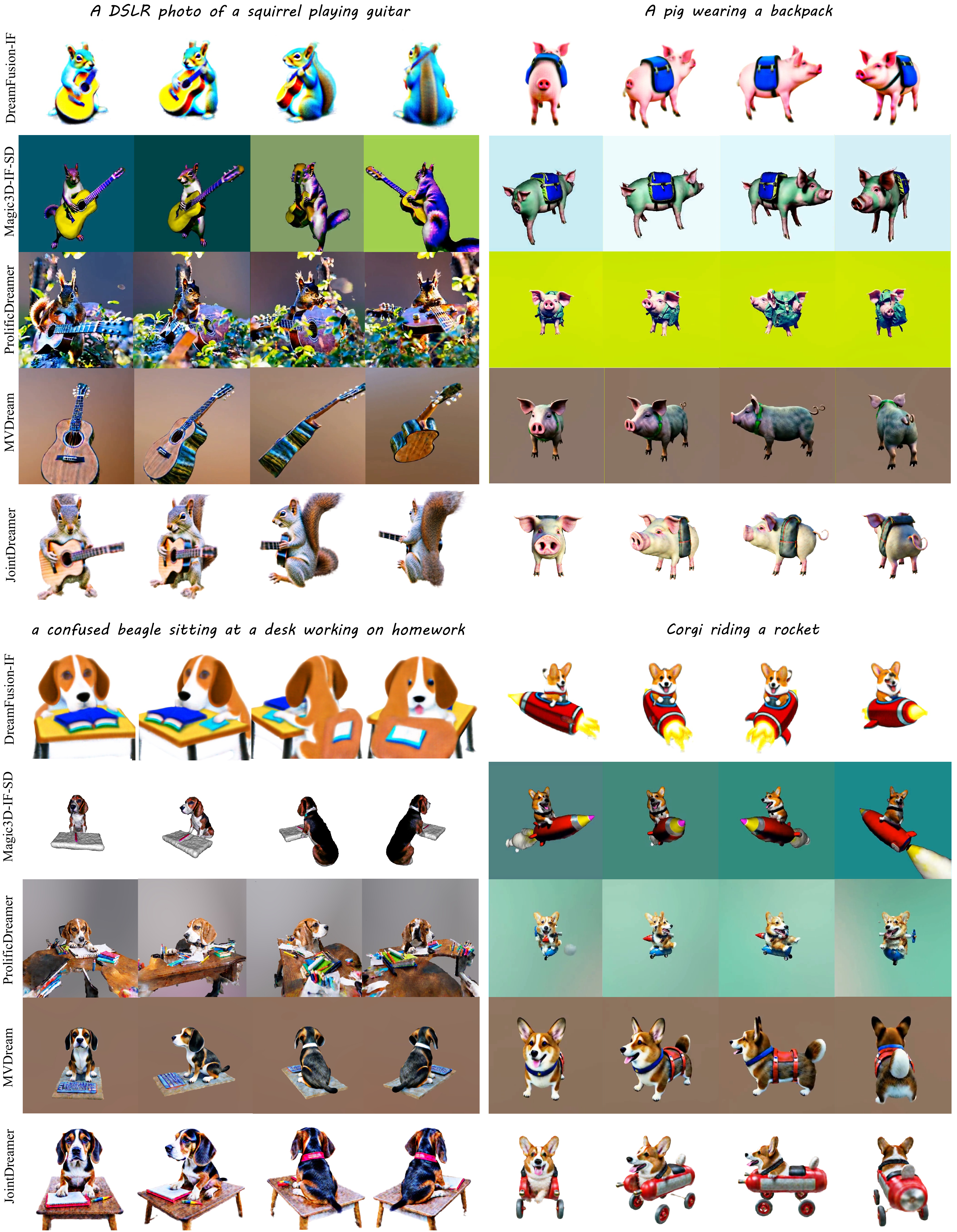}
   \vspace{-4mm}
   \caption{\textbf{More comparison of text-to-3D generation.}}
   \label{fig:quality_append1}
\end{figure}

Despite JointDreamer's impressive performance in handling detailed descriptions and multi-object combinations in long texts (as depicted in Fig. 1 of the main paper), it faces difficulties in comprehending complex relationships among objects. Specifically, it struggles to grasp relative spatial arrangements and hierarchical dependencies, as evidenced in Fig.~\ref{fig:fail}. Exploring the use of larger diffusion models, such as SDXL~\cite{podell2023sdxl}, may offer a potential solution to overcome these limitations.

\begin{figure}[ht]
  \centering
   \includegraphics[width=1\linewidth]{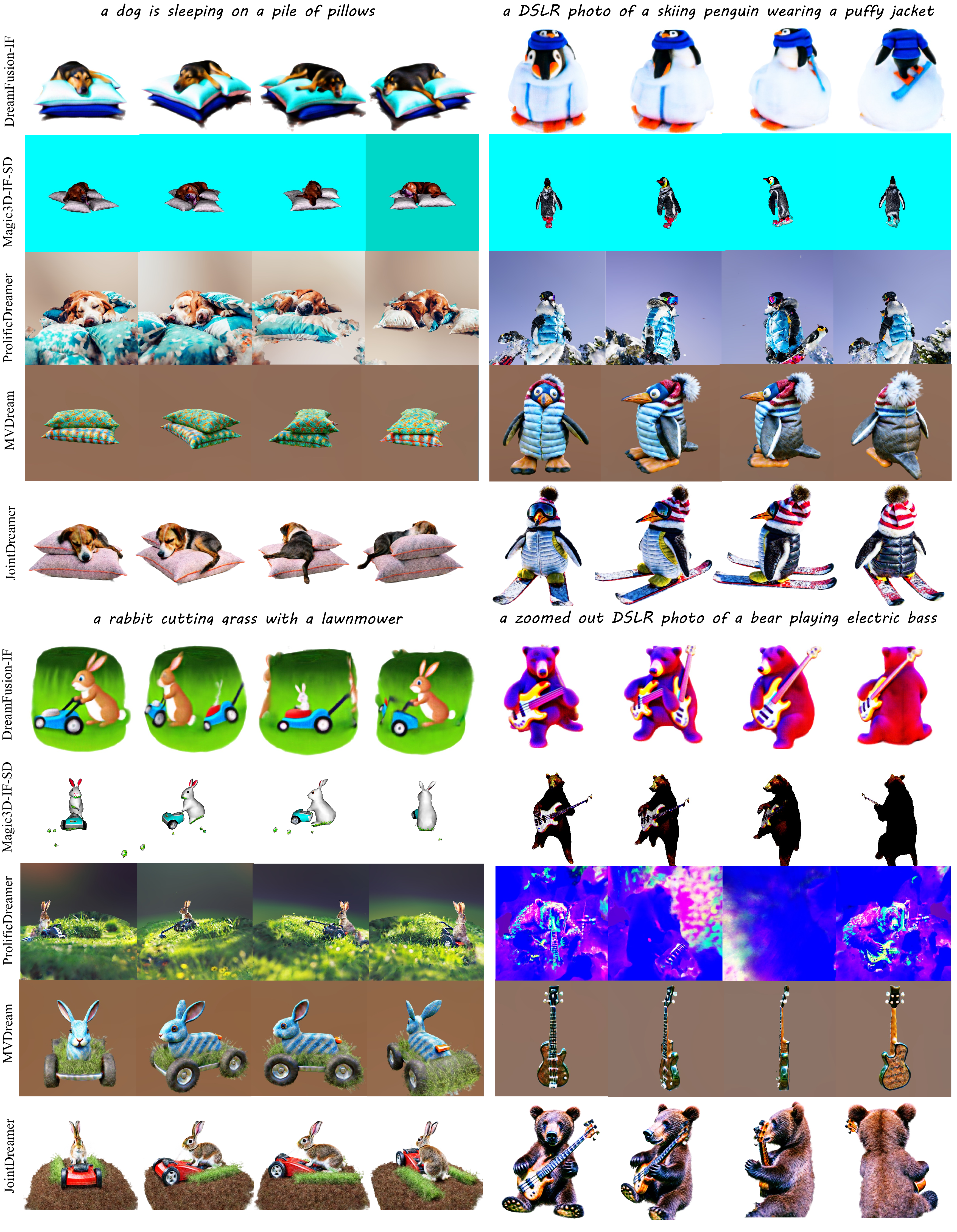}
   \vspace{-4mm}

   \caption{\textbf{More comparison of text-to-3D generation.}}
   \label{fig:quality_append2}
\end{figure}

\begin{figure}[t]
  \centering
   \includegraphics[width=1\linewidth]{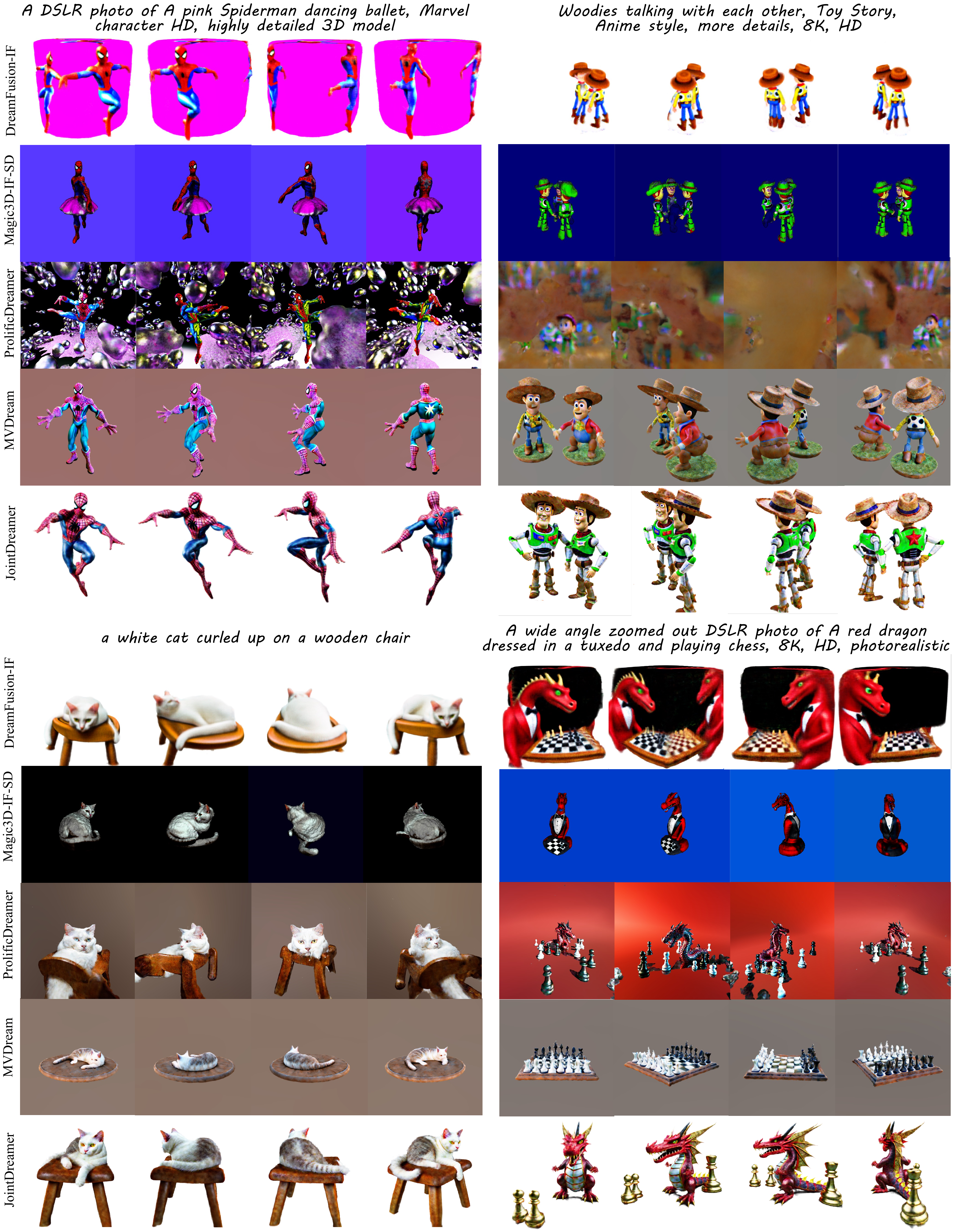}
   \vspace{-4mm}

   \caption{\textbf{More comparison of text-to-3D generation.}}
   \label{fig:quality_append3}
\end{figure}

\begin{figure}[ht]
  \centering
   \includegraphics[width=1\linewidth]{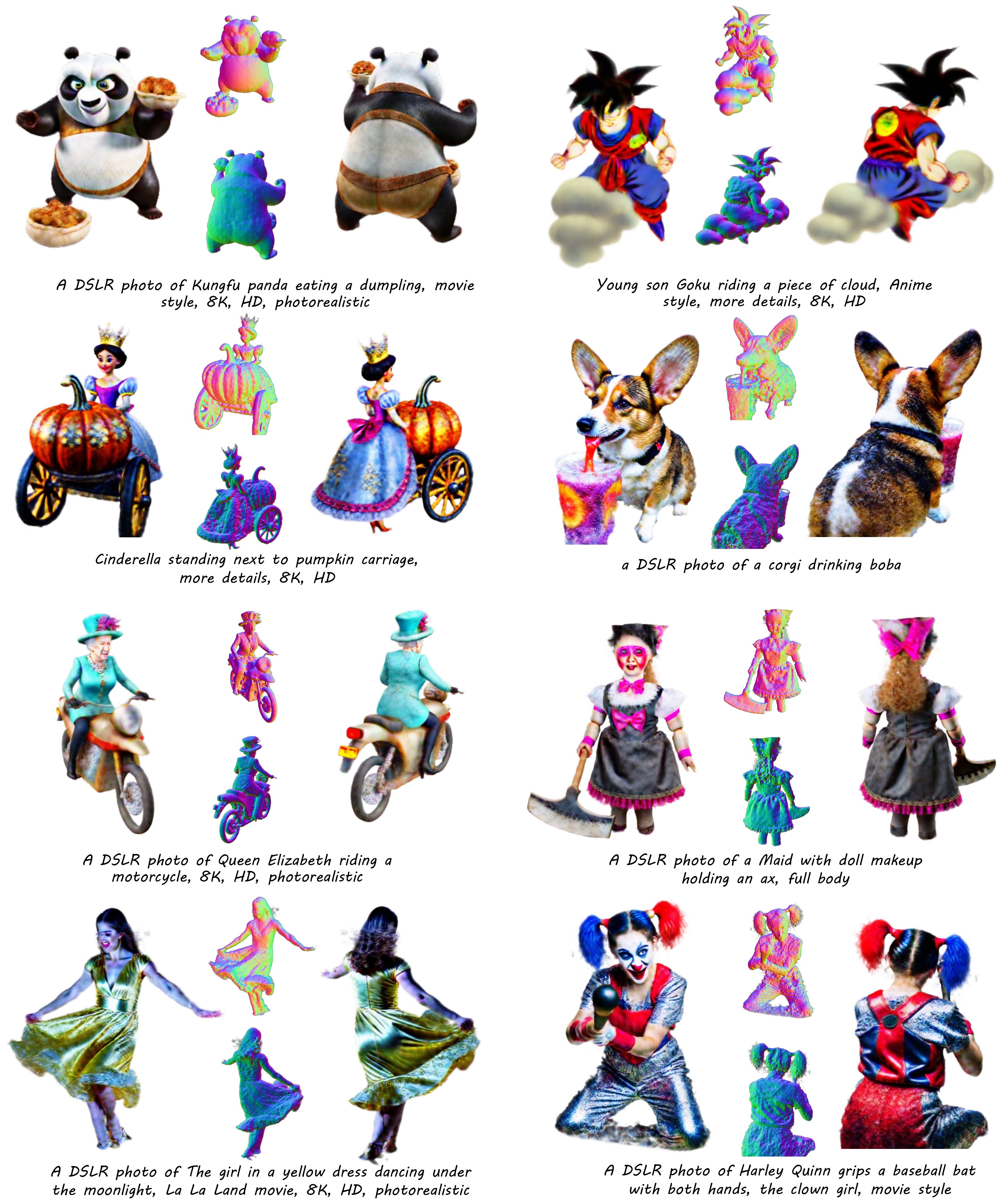}
   \vspace{-4mm}

   \caption{\textbf{More results of JointDreamer.}}
   \label{fig:results_append}
\end{figure}

\section{Additional Results of JointDreamer}\label{sec-results}

We present more comparisons of text-to-3D generation as shown in Fig.~\ref{fig:quality_append1},~\ref{fig:quality_append2} and~\ref{fig:quality_append3}. The results indicate that JointDreamer outperforms current text-to-3D generation methods regarding generation fidelity, geometric consistency, and text congruence. This further validates the effectiveness and generalization of the proposed JSD.
We also provide more images and normal maps from additional generated results in Fig.~\ref{fig:results_append}, demonstrating the generalizability of JointDreamer with arbitrary textual descriptions.

\section{Janus Prompts.}\label{sec-prompt}
Our list of 16 Janus prompts is shown below:

  "a blue jay standing on a large basket of rainbow macarons",
  
  "a confused beagle sitting at a desk working on homework",
  
  "Albert Einstein with grey suit is riding a moto",
  
  "a panda rowing a boat in a pond",
  
  "a wide angle zoomed out DSLR photo of a skiing penguin wearing a puffy jacket",
  
  "a zoomed out DSLR photo of a baby monkey riding on a pig",
  
  "a zoomed out DSLR photo of a fox working on a jigsaw puzzle",
  
  "a DSLR photo of a pigeon reading a book",
  
  "a DSLR photo of a cat lying on its side 
  
  batting at a ball of yarn"

  "A crocodile playing a drum set"
  
  "a rabbit cutting grass with a lawnmower",
  
  "A red dragon dressed in a tuxedo and playing chess",
  
  "a zoomed out DSLR photo of a bear playing electric bass", 
  
  "A bald eagle carved out of wood, more detail",
  
  "A pig wearing back pack".
  
  "a lemur drinking boba".

%% file: sub_fig/wrap_clstrain.tex
\begin{wrapfigure}{r}{0.5\textwidth}
\vspace{-4mm}
  \centering
   \includegraphics[width=1\linewidth]{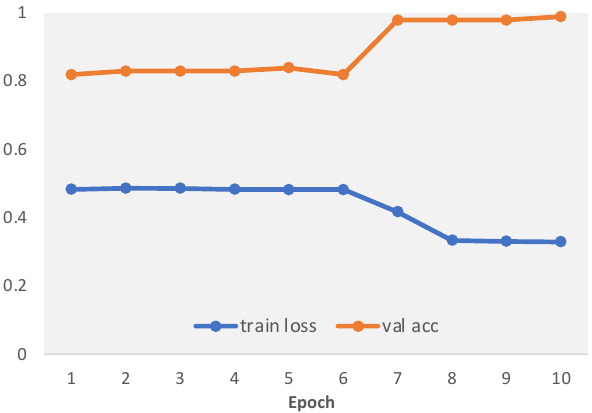}
   \vspace{-5mm}
 \caption{Training loss and validation accuracy curves of the proposed Binary Classification Model.}\label{fig:clstrain}
      \vspace{-5mm}
\end{wrapfigure}